\documentclass{article}

\usepackage{microtype}
\usepackage{graphicx}
\usepackage{subcaption}
\usepackage{booktabs} 

\usepackage{hyperref}

\usepackage[accepted]{icml2026}

\usepackage{amsmath,amsfonts,bm}

\def\eqref#1{equation~\ref{#1}}

\def\1{\bm{1}}

\def\rvk{{\mathbf{k}}}

\def\rvm{{\mathbf{m}}}
\def\rvn{{\mathbf{n}}}

\def\rvq{{\mathbf{q}}}

\def\rmK{{\mathbf{K}}}

\def\rmP{{\mathbf{P}}}
\def\rmQ{{\mathbf{Q}}}
\def\rmR{{\mathbf{R}}}

\def\rmV{{\mathbf{V}}}

\DeclareMathAlphabet{\mathsfit}{\encodingdefault}{\sfdefault}{m}{sl}
\SetMathAlphabet{\mathsfit}{bold}{\encodingdefault}{\sfdefault}{bx}{n}

\usepackage{balance}
\usepackage{xspace}
\usepackage{tikz}
\usepackage{pgfplots}
\usepackage{tabularx}
\usepackage{threeparttable}
\usepackage{pifont}
\newcommand{\ourmethod}{{LazyAttention}\xspace}
\newcommand{\codebaseurl}{\url{https://github.com/illinoisdata/lazy-attention}}

\newcommand{\blue}[1]{#1}

\newcommand{\nop}[1]{}

\usetikzlibrary{arrows.meta,positioning,calc,shapes.geometric,decorations.text,patterns,patterns.meta,fit,backgrounds,chains,shadows,math,circuits.ee.IEC,decorations.pathmorphing, decorations.pathreplacing, decorations.shapes}

\definecolor{colorset1}{RGB}{169, 194, 225}
\definecolor{colorset2}{RGB}{61, 104, 168}
\definecolor{colorset3}{RGB}{31, 60, 104}
\definecolor{colorset4}{RGB}{35, 31, 32}
\definecolor{colorset5}{RGB}{0, 0, 0}         
\definecolor{colorLazy}{RGB}{247, 168, 0}

\definecolor{barFull}{HTML}{2E64A1}
\definecolor{barLazy}{HTML}{F3B404}
\definecolor{barLightBlue}{HTML}{A9CCE3}

\definecolor{cAmain}{HTML}{1f77b4}  
\colorlet{cAlight}{cAmain!25}
\definecolor{cBmain}{HTML}{ff7f0e}  
\colorlet{cBlight}{cBmain!25}
\definecolor{cCmain}{HTML}{2ca02c}  
\colorlet{cClight}{cCmain!25}
\definecolor{cDmain}{HTML}{d62728}  
\colorlet{cDlight}{cDmain!25}
\definecolor{cEmain}{HTML}{9467bd}  
\colorlet{cElight}{cEmain!25}
\definecolor{cFmain}{HTML}{8c564b}  
\colorlet{cFlight}{cFmain!25}
\definecolor{cGmain}{HTML}{e377c2}  
\colorlet{cGlight}{cGmain!25}
\definecolor{cZmain}{HTML}{030303}  
\colorlet{cZlight}{cZmain!25}
\colorlet{cZlightlight}{cZmain!5}
\definecolor{cPositivemain}{HTML}{2ca02c}  
\colorlet{cPositivelight}{cPositivemain!25}
\definecolor{cNegativemain}{HTML}{d62728}  
\colorlet{cNegativelight}{cNegativemain!25}

\definecolor{MyRed}{RGB}{228,26,28}   
\definecolor{MyBlue}{RGB}{55,126,184}  
\definecolor{MyGreen}{RGB}{77,175,74}    
\definecolor{MyOrange}{RGB}{255,127,0}  
\definecolor{MyPurple}{RGB}{152,78,163}
\definecolor{MyBrown}{RGB}{166,86,40}   
\definecolor{MyPink}{RGB}{247,129,191} 
\definecolor{MyGray}{RGB}{153,153,153}

\usepackage{graphicx}
\usepackage{tikz}
\pgfplotsset{compat=newest}
\usepackage{pgfplots}
\usepackage{threeparttable}
\usepackage{listings}  
\usepackage{enumitem}
\usepackage{url}
\lstdefinestyle{mystyle}{
    language=C++, 
    basicstyle=\ttfamily\footnotesize, 
    keywordstyle=\color{blue}, 
    commentstyle=\color{gray}, 
    stringstyle=\color{red}, 
    numberstyle=\tiny\color{gray},
    numbers=left,
    stepnumber=1,
    frame=single, 
    tabsize=4,
    breaklines=true,
    captionpos=b
}
\usetikzlibrary{patterns}
\usetikzlibrary{calc,positioning,fit}
\usepgfplotslibrary{groupplots,units}

\definecolor{red1}{HTML}{C82423}
\definecolor{red2}{HTML}{FF8884}
\definecolor{red3}{HTML}{F8AC8C}
\definecolor{blue1}{HTML}{14517C}
\definecolor{blue2}{HTML}{2878B5}
\definecolor{blue3}{HTML}{9AC9DB}
\definecolor{barOrange}{HTML}{bb5555}
\definecolor{BlueColor}{HTML}{bb5555}
\definecolor{RedColor}{HTML}{bb5555}
\definecolor{OrangeColor}{HTML}{bb5555}
\definecolor{barYellow}{HTML}{ee9944}
\definecolor{barLightGreen}{HTML}{A3A847}
\definecolor{barGreen}{HTML}{5588bb}
\definecolor{GreenColor}{HTML}{5588bb}

\definecolor{vintageblack}{HTML}{484043}
\definecolor{vintageyellow}{HTML}{FFC805}
\definecolor{vintagegreen}{HTML}{38A528}
\definecolor{vintageorange}{HTML}{FF4D25}

\definecolor{OursColor}{HTML}{C62E2E}
\definecolor{PandasColor}{HTML}{0081a7}
\definecolor{ModinColor}{HTML}{f4a261}
\definecolor{PolarsColor}{HTML}{6D6D6D}
\definecolor{Lightgrey}{HTML}{dadada}

\newcommand*{\circled}[1]{\lower.7ex\hbox{\tikz\draw (0pt, 0pt)%
    circle (.5em) node {\makebox[1em][c]{\small #1}};}}

\usepackage{amsmath}
\usepackage{amssymb}
\usepackage{mathtools}
\usepackage{amsthm}

\usepackage[capitalize,noabbrev]{cleveref}

\theoremstyle{plain}
\newtheorem{theorem}{Theorem}[section]

\newtheorem{example}[theorem]{Example}
\newtheorem{fact}[theorem]{Fact}
\theoremstyle{definition}

\theoremstyle{remark}

\icmltitlerunning{LazyAttention: Efficient RAG with Deferred Positional Encoding}

\begin{document}

\twocolumn[
  \icmltitle{LazyAttention: Efficient Retrieval-Augmented Generation with\\ Deferred Positional Encoding}



  \icmlsetsymbol{equal}{*}

  \begin{icmlauthorlist}
    \icmlauthor{Haocheng Xia}{uiuc}
    \icmlauthor{Mihir Pamnani}{n}
    \icmlauthor{Hanxi Fang}{amz}
    \icmlauthor{Supawit Chockchowwat}{goog}
    \icmlauthor{Yongjoo Park}{uiuc}
  \end{icmlauthorlist}

  \icmlaffiliation{uiuc}{Siebel School of Computing and Data Science, University of Illinois Urbana-Champaign}
  \icmlaffiliation{goog}{Google}
  \icmlaffiliation{amz}{Amazon}
  \icmlaffiliation{n}{Nexla}

  \icmlcorrespondingauthor{Yongjoo Park}{yongjoo@illinois.edu}

  \icmlkeywords{Machine Learning, ICML}

  \vskip 0.3in
]



\printAffiliationsAndNotice{}  

\begin{abstract}
  Key-value (KV) caching accelerates inference of large language models (LLMs) by reusing past computations for generated tokens. Its importance becomes even greater in long-context applications such as retrieval-augmented generation (RAG) and in-context learning (ICL). However, conventional KV caching embeds positional information directly into the cache, limiting its reusability. Existing solutions either restrict reuse to prefixes or require expensive memory materialization for positional re-encoding.
  We introduce \ourmethod, a novel attention mechanism that {kernelizes} deferred positional encoding to enable {zero-copy, position-agnostic} KV reuse. By adjusting positional encoding within attention kernels \textit{on-the-fly}, \ourmethod resolves the materialization bottleneck, allowing a single physical KV copy to serve multiple logical requests at arbitrary positions.
  Leveraging attention kernels tailored for prefilling and decoding, our system achieves significant efficiency improvements: under skewed document distributions, it reduces time-to-first-token (TTFT) by 1.37$\times$ and increases inference throughput by 1.40$\times$ compared to the state-of-the-art Block-Attention, while maintaining comparable output quality. 
\end{abstract}

\section{Introduction}\label{sec:Introduction}

Retrieval-augmented generation (RAG) greatly improves the quality and timeliness of responses by enriching user queries with external data~\citep{Lewis2020RAG,DBLP:conf/icml/GuuLTPC20,DBLP:conf/acm/AsaiMZC23,DBLP:journals/tacl/RamLDMSLS23,gao2024empowering,kalra-etal-2024-hypa,LangChain2024,xiong-etal-2024-benchmarking}. However, processing this external data remains a major bottleneck for achieving low-latency RAG, since answer generation can only begin once the data has been fully processed. This step scales poorly—the computational complexity grows quadratically with input length~\citep{jiang2024minference,hu2025epic,DBLP:conf/icml/TangZZXKH24,DBLP:conf/icml/AcharyaJG25,DBLP:conf/icml/ZhangXHWX0C25}.
The overhead will further worsen as modern models support increasingly longer context windows~\citep{gemini}. While each retrieved document must be processed at least once, its results could be stored in a form that enables more flexible reuse. This motivates a careful investigation into \emph{reusable components}.

Existing inference approaches use conventional key-value (KV) cache~\citep{DBLP:journals/corr/abs-2407-00079} as reusable components, but remain ineffective due to their \emph{position-awareness}. That is, cached values are reusable only if the associated data (e.g., a document) appears in the same position as before.
This restriction is considered by earlier caching techniques,
which accordingly focus on the identification of exactly matching document sequences~\citep{DBLP:conf/sosp/KwonLZ0ZY0ZS23, DBLP:conf/mlsys/GimCLSK024, DBLP:journals/corr/abs-2404-12457}.
However, the chance of observing an exact match is lower than that of encountering individual documents.
Recent works~\citep{lu-etal-2025-turborag,ma2025blockattention} show that KV cache can be reused even for the documents appearing in different positions if their positional information is re-encoded.
This enhances reusability but is still memory-inefficient because position re-encoding requires duplicating the KV cache.
In-place updates can cause race conditions and incorrect outcomes if two prompts in the same batch share the same data.
This limitation---position-awareness---makes caching ineffective.
To illustrate, suppose document popularity follows a Zipf distribution and each document may appear in any of $D$ prompt positions. With a cache budget of $C$ physical KV entries, a position-agnostic cache can store the top $C$ documents, whereas a position-aware cache spends entries on positional variants and covers only about the top $\lfloor C/D \rfloor$ documents. The resulting hit-ratio gap is
\(
\frac{\sum_{i=1}^{C} i^{-\alpha}}
     {\sum_{i=1}^{\lfloor C/D \rfloor} i^{-\alpha}},
\)
which is $2.86\times$ for $D=20$ and $C=100$ under a moderately skewed Zipf distribution; \Cref{sec:Preliminaries} gives the full analysis.

In this paper, we argue that reusable components should be \textit{position-agnostic} to enable effective caching under limited memory. On modern GPUs, high-bandwidth memory (HBM) remains a scarce resource for LLM serving, as KV cache size grows with both context length and the number of concurrent requests. Existing memory-saving techniques---including KV compression~\citep{kvcompress1,minicache}, grouped-query attention~\citep{ainslie2023gqa}, and latent KV representations~\citep{DeepSeekV2}---reduce the size of individual KV entries. However, they do not address the capacity wasted on storing position-specific variants of the same reusable content.
A position-agnostic mechanism removes this redundancy: full coverage for $N$ reusable documents requires only $O(N)$ document-level KV entries rather than multiple position-specific copies per document. Under Zipf-distributed workloads, this lets the same memory budget cover more high-probability documents and translates into the distribution-dependent cache-hit improvement analyzed in Section~\ref{sec:Preliminaries}.
This would also make it feasible to preprocess and fully cache a medium-sized database using a combination of GPU HBM and host memory, enabling near-zero latency for a wide range of RAG applications. However, this possibility is nearly impossible with existing mechanisms. Positions are \emph{eagerly} embedded into KV cache for nearly all models (e.g., BERT~\citep{bert}, Llama~\citep{llama3}, DeepSeek~\citep{DeepSeekV2}) running on the latest serving systems (e.g., vLLM~\citep{vllm}, Orca~\citep{orca}, SGLang~\citep{zheng2024sglang}). Achieving {position-agnostic} reuses will therefore require a fundamentally different approach.

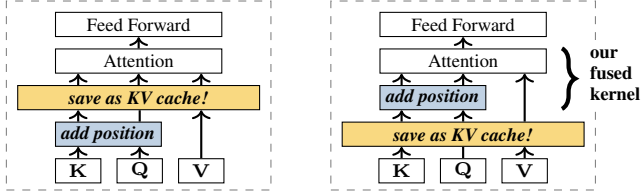
\begin{figure}[t]

\pgfmathsetmacro{\boxheight}{3.2mm}
\pgfmathsetmacro{\ygap}{0.15}

\tikzset{
  mybox/.style = {
    draw=gray, minimum width=35mm, minimum height=25mm, dashed
  },
  ff/.style = {
    draw=black, minimum width=22mm, minimum height=\boxheight,
    anchor=north, font=\scriptsize, inner ysep=0mm,
  },
  attn/.style = {
    draw=black, minimum width=22mm, minimum height=\boxheight,
    anchor=north, font=\scriptsize, inner ysep=0mm,
  },
  kv/.style = {
    draw=black, minimum width=32mm, minimum height=\boxheight,
    anchor=north, font=\scriptsize\bfseries\itshape, inner ysep=0mm,
    fill=barLazy!50!white
  },
  posen/.style = {
    draw=black, minimum width=14mm, minimum height=\boxheight,
    anchor=north west, font=\scriptsize\bfseries\itshape, inner ysep=0mm, inner xsep=0mm,
    fill=barFull!30!white,
  },
  kvq/.style = {
    draw=black, minimum width=6mm, minimum height=\boxheight,
    anchor=north west, font=\scriptsize, inner ysep=0mm,
  },
}

\centering
\begin{subfigure}[b]{0.48\linewidth}

\hspace{4mm}
\begin{tikzpicture}

\node[mybox] (T) {};

\node[ff] (ff) at ($(T.north)+(0,-\ygap)$) {Feed Forward};

\node[attn] (attn) at ($(ff.south)+(0,-\ygap)$) {Attention};

\node[kv] (kv) at ($(attn.south)+(0,-\ygap)$) {save as KV cache!};

\node[posen] (posen) at ($(kv.south west)+(0.5,-\ygap)$) {add position};

\node[kvq] (k) at ($(posen.south west)+(0,-\ygap)$) {$\rmK$};

\node[kvq,anchor=west] (q) at ($(k.east)+(0.2,0)$) {$\rmQ$};

\node[kvq,anchor=west] (v) at ($(q.east)+(0.2,0)$) {$\rmV$};

\draw[->,thick] (k) -- ($(posen.south west)!(k.north)!(posen.south east)$);
\draw[->,thick] (q) -- ($(posen.south west)!(q.north)!(posen.south east)$);
\draw[->,thick] (v) -- ($(kv.south west)!(v.north)!(kv.south east)$);
\draw[->,thick] ($(kv.north west)!(v.north)!(kv.north east)$) 
    -- ($(attn.south west)!(v.north)!(attn.south east)$);
\draw[->,thick] ($(kv.north west)!(k.north)!(kv.north east)$) 
    -- ($(attn.south west)!(k.north)!(attn.south east)$);

\draw[->,thick] ($(posen.north west)!(k.north)!(posen.north east)$) 
    -- ($(kv.south west)!(k.north)!(kv.south east)$);
\draw[-,thick] ($(posen.north west)!(q.north)!(posen.north east)$) 
    -- ($(kv.south west)!(q.north)!(kv.south east)$); 

\draw[->,thick] (kv.north) -- (attn.south);
\draw[->,thick] (attn.north) -- (ff.south);

\end{tikzpicture}

\caption{Existing positional encoding. Positional encoding (\textcolor{barFull}{blue box}) is \emph{eagerly} applied before saving (\textcolor{barLazy}{yellow box}) the computed key (K) and queries (Q).}
\end{subfigure}
\hfill
\begin{subfigure}[b]{0.48\linewidth}

\hspace{4mm}
\begin{tikzpicture}

\node[mybox] (T) {};

\node[ff] (ff) at ($(T.north)+(0,-\ygap)$) {Feed Forward};

\node[attn] (attn) at ($(ff.south)+(0,-\ygap)$) {Attention};

\node[posen] (posen) at ($(attn.south west)+(0,-\ygap)$) {add position};

\node[kv] (kv) at ($(posen.south)+(0.4,-\ygap)$) {save as KV cache!};

\node[kvq,anchor=north] (q) at ($(kv.south)+(0,-\ygap)$) {$\rmQ$};

\node[kvq,anchor=east] (k) at ($(q.west)+(-0.2,0)$) {$\rmK$};

\node[kvq,anchor=west] (v) at ($(q.east)+(0.2,0)$) {$\rmV$};

\draw[->,thick] (k) -- ($(kv.south west)!(k.north)!(kv.south east)$);
\draw[-,thick] (q) -- ($(kv.south west)!(q.north)!(kv.south east)$);
\draw[->,thick]  (v) -- ($(kv.south west)!(v.north)!(kv.south east)$);

\draw[->,thick] ($(kv.north west)!(k.north)!(kv.north east)$) 
    -- ($(posen.south west)!(k.north)!(posen.south east)$);
\draw[->,thick] ($(kv.north west)!(q.north)!(kv.north east)$) 
    -- ($(posen.south west)!(q.north)!(posen.south east)$);

\draw[->,thick] ($(posen.north west)!(k.north)!(posen.north east)$) 
    -- ($(attn.south west)!(k.north)!(attn.south east)$);
\draw[->,thick] ($(posen.north west)!(q.north)!(posen.north east)$) 
    -- ($(attn.south west)!(q.north)!(attn.south east)$);
\draw[->,thick] ($(kv.north west)!(v.north)!(kv.north east)$) 
    -- ($(attn.south west)!(v.north)!(attn.south east)$);

\draw[->,thick] (attn.north) -- (ff.south);

\draw [decorate,decoration={brace,amplitude=5pt},ultra thick] 
    ($(attn.north east)+(0.2,0)$) -- ++(0,-0.8);
\node[font=\scriptsize\bfseries,anchor=west,align=left,fill=white] 
    at ($(attn.north east)+(0.45,-0.35)$) {our\\fused\\ kernel};

\end{tikzpicture}

\caption{Our positional encoding. Positional encoding (\textcolor{barFull}{blue box}) is \emph{lazily} applied inside attention kernels and KV caching (\textcolor{barLazy}{yellow box}) logically omits positions.}
\end{subfigure}

\caption{Comparison of positional encoding strategies within a Transform block. 
Existing methods, even the state-of-the-art Block-Attention for KV cache reuse, eagerly apply positions before caching, making KV caches position-dependent and non-shareable. \ourmethod applies positions lazily inside the fused attention kernel, enabling reuse and sharing without duplication.}
\label{fig:intro}
\end{figure}

We propose \ourmethod, an attention mechanism that can enable \textit{position-agnostic} KV reuse, significantly improving cache effectiveness for RAG applications.
{While the high-level concept of decoupling rotary position embedding (RoPE) from KV storage has been explored~\citep{lu-etal-2025-turborag,ma2025blockattention}, prior attempts faced a critical \emph{memory-compute trade-off}: they either required materializing position-adjusted copies (high memory/bandwidth cost) or restricted reuse to prefixes causing more recomputation (high computation cost).
\ourmethod resolves this trade-off by \emph{kernelizing} the RoPE-decoupling technique.}
The key idea shown in Figure~\ref{fig:intro} is to develop an alternative form of KV cache that can be shared among multiple prompts \emph{without} approximating attention computation. \ourmethod achieves this by \emph{deferring}, rather than omitting, positional encoding until the final stage---when the attention is computed. At that point, the kernel dynamically encodes positions on chip by considering relative distances between each query-key pair, requiring only a few additional kernel variables.
Importantly, positional encoding in \ourmethod is only \textit{transient}, lasting only for the brief period of attention computation.
This design completely avoids additional data materialization in HBM. Despite the difference, \ourmethod's attention is mathematically identical to existing methods that duplicate KV caches for position re-encoding~\citep{lu-etal-2025-turborag,ma2025blockattention}.
As a result, it produces the same attention scores and generated answers.

\ourmethod is implemented based on vLLM~\citep{vllm} with two custom attention kernels for prefilling and decoding in Triton~\citep{triton}. Designing separate kernels is non-trivial, as prefilling is typically compute-bound while decoding is memory-bandwidth-bound. To address these distinct bottlenecks, our kernels are carefully optimized to minimize extra computation in prefilling and extra I/O in decoding. Furthermore, our custom attention mechanism integrates positional encoding directly into the key-value matrix multiplications. Despite this added logic, the runtime overhead remains around $0.2\%$ in practice, while substantially improving cache effectiveness and reducing latency.

We evaluated \ourmethod against diverse baselines, including CacheBlend~\citep{DBLP:conf/eurosys/cacheblend}, Prompt Cache~\citep{DBLP:conf/mlsys/GimCLSK024}, and Block-Attention~\citep{ma2025blockattention}, on standard RAG benchmarks, demonstrating consistent improvements in latency and throughput empirically. Our approach is particularly effective in common RAG scenarios with skewed document access patterns, where hot documents are frequently reused: compared to the state-of-the-art Block Attention, \ourmethod reduces time-to-first-token (TTFT) by $1.37\times$ and increases inference throughput by $1.40\times$, while maintaining similar output quality. While our primary evaluation focuses on RAG, \ourmethod also benefits any workload where text chunks recur across requests, such as few-shot in-context learning (Appendix~\ref{appx:beyond_rag}) and parallel hypothesis agents~\cite{fang2025large}. Our contribution can be summarized as follows.

\begin{itemize}[leftmargin=*,itemsep=-1mm,topsep=-1mm]
\item {We propose a novel attention mechanism that kernelizes the RoPE-decoupling technique to enable zero-copy, position-agnostic KV reuse. By deferring positional encoding to a transient step within the attention kernel, our method resolves the memory--compute trade-off that previously hindered the scaling of arbitrary-position reuse.}
\item We demonstrate that this position-agnostic design significantly increases the cache hit ratio. A single cached document entry can be shared across all requests, regardless of its position, maximizing cache efficiency under memory constraints. For skewed access patterns, the hit ratio improves by 7.5$\times$ compared to prefix caching.
\item We provide highly optimized Triton kernels~\footnote{\codebaseurl} for both prefilling and decoding that implement our mechanism with negligible overhead (around 0.2\%) even for long-context inputs, translating our architectural improvements into substantial practical gains in end-to-end throughput and latency for RAG workloads.
\end{itemize}

\section{Background and Motivation}\label{sec:Preliminaries}

In this section, we briefly review the literature of RAG and KV cache reuse, then discuss why existing techniques struggle with efficiency under dynamic contexts.  


Retrieval-augmented generation (RAG)~\citep{LangChain2024,DBLP:conf/icml/GuuLTPC20,Lewis2020RAG,CAG,pbemeets} augments LLMs with external knowledge. Given a query \(\mathcal{Q}\), a retriever selects top-\(N\) relevant documents \(\mathcal{D}=\{d_1,\ldots,d_N\}\), which the LLM conditions on when producing an answer. The most common approach is simple concatenation of retrieved documents and the query~\citep{DBLP:conf/eacl/IzacardG21}, though more complex schemes exist~\citep{DBLP:conf/icml/BorgeaudMHCRM0L22}.  

RAG enables flexible knowledge grounding but also introduces variability: newly collected documents may need to be inserted while irrelevant or low-quality documents may need to be removed across queries~\citep{DBLP:conf/acl/LiRZWLVYK23,DBLP:conf/iclr/YoranWRB24,DBLP:conf/sigir/CuconasuTSFCMTS24}. Such changes alter the prefix, forcing standard attention to re-run the costly prefilling step. This motivates techniques for reusing previously computed KV caches beyond the prefix to support dynamic contexts efficiently.  

Existing methods, including Prompt Cache~\citep{DBLP:conf/mlsys/GimCLSK024}, CacheBlend~\citep{DBLP:conf/eurosys/cacheblend}, EPIC~\cite{hu2025epic}, TurboRAG~\cite{lu-etal-2025-turborag}, and Block-Attention~\citep{ma2025blockattention}, all rely on re-encoding positions. While effective to some extent, this coupling between cache and position leads to inefficiency: each distinct document position can consume a separate KV-cache copy. In short, existing methods either suffer from low reusability or exhaust memory with redundant cache copies. 

\paragraph{Hit-ratio impact under limited cache capacity}
We now quantify why position-awareness directly lowers cache hit ratio. Assume reusable documents follow a Zipf distribution, where the $i$-th most popular document has probability proportional to $i^{-\alpha}$. If each document may appear in any of $D$ prompt positions and the cache stores $C$ physical KV entries, a position-agnostic cache stores one entry per document and caches the top $C$ documents:
\begin{equation}
H_{\mathrm{agnostic}}(C) = \frac{\sum_{i=1}^{C} i^{-\alpha}}{\sum_{j=1}^{N} j^{-\alpha}},
\end{equation}
where $N$ is the number of reusable documents. A position-aware cache must instead spend capacity on positional variants. If each cached document needs up to $D$ position-specific copies, the same budget covers only about $\lfloor C/D \rfloor$ unique documents:
\begin{equation}
H_{\mathrm{aware}}(C,D) \approx \frac{\sum_{i=1}^{\lfloor C/D \rfloor} i^{-\alpha}}{\sum_{j=1}^{N} j^{-\alpha}}.
\end{equation}
The resulting hit-ratio advantage is
\begin{equation}
\frac{H_{\mathrm{agnostic}}(C)}{H_{\mathrm{aware}}(C,D)} \approx
\frac{\sum_{i=1}^{C} i^{-\alpha}}{\sum_{i=1}^{\lfloor C/D \rfloor} i^{-\alpha}}.
\label{eq:hit-ratio-analysis}
\end{equation}
For example, when documents may appear in $D=20$ positions and the cache can store $C=100$ KV entries, \Cref{eq:hit-ratio-analysis} gives a $2.86\times$ hit-ratio advantage under a moderately skewed Zipf distribution. The gap arises under the same physical memory budget because position-aware caching spends entries on duplicate positional variants, while position-agnostic caching covers more unique documents. This raises the central question: \textit{can we achieve high cache hit ratios under limited memory without materializing position-specific KV copies?}

Relative positional encodings such as RoPE~\citep{RoPE} suggest a possible remedy: keys can be rotated on-the-fly to realign with the new positions of reused documents. However, applying rotations dynamically during decoding requires recomputing large key matrices for every token, inducing prohibitive overhead that violates the strict latency constraints of real-time inference.

\vspace{0.8em}
\section{\textsc{LazyAttention}: Algorithm and Analysis}

In this section, we show how the positional information of reused documents can be adjusted \textit{during} the attention calculation. Then we analyze the cost of deferred positional encoding for prefilling and decoding, respectively. Inspired by the analysis, we present how to integrate \ourmethod with FlashAttention~\citep{flashattention,flashattention2,flashattention3} seamlessly to achieve efficient computation. 

\paragraph{Problem statement} Motivated by the limitations of existing approaches and the constraints of GPU memory, the problem is to develop a method that efficiently manages and manipulates the positional information of documents in KV caches \textit{on the fly} during attention computation, without incurring high recomputation or extensive copying costs. The method should correctly handle the global positional alignment required by the request and support scenarios involving multiple requests accessing the same document at different global offsets within a batch.

\subsection{Deferred Positional Encoding}

In the standard Transformer architectures~\cite{transformer}, positional encoding is applied before attention computation, forcing explicit duplication of the KV cache whenever document positions differ. Our approach logically \emph{defers} positional encoding until the attention computation step, ensuring the KV cache remains position-agnostic.  

\paragraph{Standard attention}  
In the standard attention formulation, given query, key, and value matrices, $\rmQ$, $\rmK$, and $\rmV$, the attention output is computed as follows,  
\begin{equation}
    \operatorname{Attention}(\rmQ,\rmK,\rmV) = 
    \operatorname{softmax}\!\left(\frac{\rmQ \rmK^\top}{\sqrt{d_k}}\right)\rmV,
    \label{Eqn:standard_attention}
\end{equation}
where $d_k$ is the dimensionality of $\rmK$.  

Positional encoding allows the model to exploit sequence order when computing attention scores, e.g., allocate more attention to the recent context. A variety of methods exist---ranging from absolute to relative encodings such as XLNet-style~\cite{xlnet}, rotary positional embedding (RoPE)~\cite{RoPE}, and interleaved rotary position embedding~\cite{llama4}. Among these, {RoPE} has become the most widely adopted due to its simplicity and effectiveness. In this paper, we focus on the original RoPE and its variants~\cite{llama3.1,rope-scaling}. With RoPE, Equation~\ref{Eqn:standard_attention} becomes:  
\begin{equation}
    \operatorname{Attention}(\rmQ,\rmK,\rmV) = 
    \operatorname{softmax}\!\left(\frac{f(\rmQ,\rvm) f(\rmK,\rvn)^\top}{\sqrt{d_k}}\right)\rmV,\nonumber
    \label{Eqn:rope_attention}
\end{equation}
where $f$ denotes the positional encoding function, and $\rvm,\rvn$ represent the positional indices of the tokens in $\rmQ$ and $\rmK$.  

\paragraph{Deferred encoding}  
In our method, the query, key, and value matrices are first computed from the input sequence without positional information. Positional encoding is applied only \emph{during} attention computation, after retrieving entries from the KV cache. This means the cache is regarded as purely content-based keys and values, which can be reused across arbitrary positions. The positional adjustments are applied at runtime, yielding both mathematical correctness and significantly improved cache efficiency, as shown in the following example.  

\begin{example}\label{Example:rotation}

    Suppose a query $\mathcal{Q}$ encodes its positional index relative to the document length. Consider two documents, $d_1$ and $d_2$, with independently generated KV caches $\mathcal{C}_1$ and $\mathcal{C}_2$, both indexed from position $0$. To reuse $\mathcal{C}_2$ for $Q$ immediately following the processing of $d_1$, we rotate $Q$ backward by $|d_1|$ before the attention computation. This adjustment ensures positional consistency by aligning the query's phase with the pre-computed cache.

\end{example}

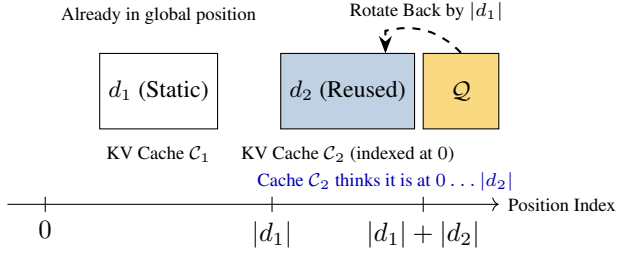
\begin{figure}
    \centering
\begin{tikzpicture}[
    node distance=0cm,
    block/.style={rectangle, draw, minimum height=1cm, minimum width=1cm, font=\small},
    cache/.style={fill=barFull!30!white},
    query/.style={fill=barLazy!50!white},
    arrow/.style={-Stealth, thick}
]

\draw[->] (-0.5,-1.5) -- (6,-1.5) node[right,font=\scriptsize] {Position Index};
\foreach \x/\label in {0/0, 3/|d_1|, 5/|d_1|+|d_2|}
    \draw (\x, -1.4) -- (\x, -1.6) node[below] {$\label$};

\node[block] (d1) at (1.5,0) {$d_1$ (Static)};
\node[below=0.1cm of d1, font=\scriptsize] {KV Cache $\mathcal{C}_1$};

\node[block, cache] (d2) at (4,0) {$d_2$ (Reused)};
\node[below=0.1cm of d2, font=\scriptsize] {KV Cache $\mathcal{C}_2$ (indexed at $0$)};

\node[block, query, minimum width=1cm] (q) at (5.5,0) {$\mathcal{Q}$};

\draw[arrow, bend right=30, dashed] (q.north) to node[above, font=\scriptsize] {Rotate Back by $|d_1|$} (4.5, 0.8) -- (4.5, 0.5);

\node[align=left, font=\scriptsize] at (1.5, 1) {Already in global position};
\node[align=left, font=\scriptsize, color=blue!70!black] at (4.5, -1.2) {Cache $\mathcal{C}_2$ thinks it is at $0 \dots |d_2|$};

\end{tikzpicture}
\caption{Illustration of Example~\ref{Example:rotation}.}
    \label{fig:placeholder}
\end{figure}

\begin{figure*}[t]

\tikzset{
  qv/.style = {
    draw=black, minimum width=4mm, minimum height=12mm, font=\scriptsize
  },
  kk/.style = {
    draw=black, minimum width=16mm, minimum height=3mm, font=\scriptsize, inner ysep=0.5mm,
  },
  mylabel/.style={
    font=\scriptsize, inner sep=0.2mm, fill=white, fill opacity=0.9
  }
}

\centering
\begin{subfigure}[b]{0.31\linewidth}

\centering
\begin{tikzpicture}
\node[draw=black, minimum width=16mm, minimum height=12mm] (A) {};

\node[qv, fill=barFull!40!white, anchor=east,opacity=0] (R) at ($(A.west)+(-0.2,0)$) {R};
\node[qv, anchor=east] (Q) at ($(R.west)+(0.2,0)$) {};
\node[qv, anchor=west] (V) at ($(A.east)+(0.2,0)$) {};

\node[kk, anchor=south, fill=barFull!40!white] (R2) at ($(A.north)+(0,0.2)$) {R $\times$ 2};
\node[kk, anchor=south] (K) at ($(R2.north)+(0,0.2)$) {};


\node[mylabel] at (Q) {Q};
\node[mylabel] at (V) {V};
\node[mylabel] at (K) {K};

\node[font=\scriptsize, minimum width=8mm, minimum height=6mm, draw=black,fill=barLazy!40!white] 
    at ($(A.north west)+(0.6,-0.75)$) {A};

\coordinate (Qs) at ($(Q.north east)+(0,-0.75)$);
\draw[very thick, ->] (Qs) -- ++(0.6,0);

\coordinate (Ks) at ($(K.south west)+(0.6,0)$);
\draw[very thick, ->] (Ks) -- ++(0,-0.2);
\draw[very thick, ->] ($(Ks)+(0,-0.5)$) -- ++(0,-0.7);

\coordinate (Vs) at ($(V.north west)+(0,-0.45)$);
\draw[very thick, ->] (Vs) -- ++(-0.5,0) -- ++(0,-0.3) -- ++(-0.3,0);

\end{tikzpicture}

\caption{Existing: keys are rotated (R) twice before attention (A)}
\end{subfigure}
\hfill
\begin{subfigure}[b]{0.31\linewidth}

\centering
\begin{tikzpicture}
\node[draw=black, minimum width=16mm, minimum height=12mm] (A) {};

\node[qv, anchor=north east, minimum height=5.5mm] (Q1) at ($(R.north west)+(0.2,0)$) {$\text{Q}_1$};
\node[qv, anchor=south east, minimum height=5.5mm] (Q2) at ($(R.south west)+(0.2,0)$) {$\text{Q}_2$};
\node[qv, anchor=west] (V) at ($(A.east)+(0.2,0)$) {};

\node[kk, anchor=south, opacity=0] (R2) at ($(A.north)+(0,0.2)$) {R};

\node[kk, anchor=south west, minimum width=7.5mm] (K1) at ($(R2.north west)+(0,0.2)$) {$\text{K}_1$};
\node[kk, anchor=south east, minimum width=7.5mm] (K2) at ($(R2.north east)+(0,0.2)$) {$\text{K}_2$};


\node[mylabel] at (V) {V};

\node[font=\scriptsize, minimum width=8mm, minimum height=6mm, draw=black,,fill=barLazy!40!white] (AA)
    at ($(A.north west)+(0.6,-0.75)$) {};

\node[anchor=north,font=\scriptsize,fill=barFull!40!white,inner xsep=3mm,inner ysep=0.5mm] at (AA.north) {R};
\node[anchor=south,font=\scriptsize] 
    at ($(AA.south)+(-0.02,0)$) {+A};

\coordinate (Qs) at ($(Q1.north east)+(0,-0.75)$);
\draw[very thick, ->] (Q1.east) -- ++(0.3,0) -- ++(0,-0.3)-- ++(0.3,0)  ;
\draw[very thick, ->] (Q2.east) -- ++(0.6,0);

\draw[very thick, ->] (K1.south) -- ++(0,-1.2);
\draw[very thick, ->] (K2.south) -- ++(0,-0.6) -- ++(-0.5,0) -- ++(0,-0.6);

\coordinate (Vs) at ($(V.north west)+(0,-0.45)$);
\draw[very thick, ->] (Vs) -- ++(-0.5,0) -- ++(0,-0.3) -- ++(-0.3,0);
\end{tikzpicture}

\caption{Our prefilling: keys are rotated only once inside the fused attention kernel}
\end{subfigure}
\hfill
\begin{subfigure}[b]{0.31\linewidth}

\centering
\begin{tikzpicture}
\node[draw=black, minimum width=16mm, minimum height=12mm] (A) {};

\node[qv, fill=barFull!40!white, anchor=east, opacity=0] (R) at ($(A.west)+(0.2,0)$) {R};
\node[qv, anchor=east] (Q) at ($(R.west)+(-0.2,0)$) {};
\node[qv, anchor=west] (V) at ($(A.east)+(0.2,0)$) {};

\node[kk, anchor=south, fill=barFull!40!white, opacity=0] (R2) at ($(A.north)+(0,0.2)$) {R};
\node[kk, anchor=south] (K) at ($(R2.north)+(0,0.2)$) {};


\node[mylabel] at (Q) {Q};
\node[mylabel] at (V) {V};
\node[mylabel] at (K) {K};

\node[font=\scriptsize, minimum width=8mm, minimum height=6mm, draw=black,fill=barLazy!40!white]  (AA)
    at ($(A.north west)+(0.6,-0.75)$) {};

\coordinate (Qs) at ($(Q.north east)+(0,-0.75)$);
\draw[very thick, ->] (Qs) -- (AA.west);

\coordinate (Ks) at ($(K.south west)+(0.6,0)$);
\draw[very thick, ->] (Ks) -- (AA.north);

\coordinate (Vs) at ($(V.north west)+(0,-0.45)$);
\draw[very thick, ->] (Vs) -- ++(-0.5,0) -- ++(0,-0.3) -- ++(-0.3,0);

\node[anchor=west,font=\scriptsize,fill=barFull!40!white,inner xsep=1mm,inner ysep=2mm] at (AA.west) {R};
\node[anchor=east,font=\scriptsize] 
    at ($(AA.east)+(-0.02,0)$) {+A};

\end{tikzpicture}

\caption{Our decoding: query is rotated inside the fused attention kernel}
\end{subfigure}

\caption{Comparison of Rotary Embedding (R) placement in tiled attention.
\textit{(a)} Naive deferred encoding: Rotates keys twice (resetting to zero, then shifting to target) before attention (A), incurring high computational overhead.
\textit{(b)} \ourmethod (prefilling): Rotates only the keys inside the attention kernel on the fly, allowing Keys/Values to be read as-is to avoid materialization overhead ($\text{K}_1/\text{K}_2$ denote the RoPE half-dimensions). The rotary vector is loaded from HBM to avoid extra computation.
\textit{(c)} \ourmethod (decoding): Fuses rotation directly into the attention inner loop, applying it to Q on-the-fly via relative offsets, ensuring the KV cache remains position-agnostic with negligible cost. The rotary vector is computed on the fly to save bandwidth.}
\label{fig:tiling}
\end{figure*}
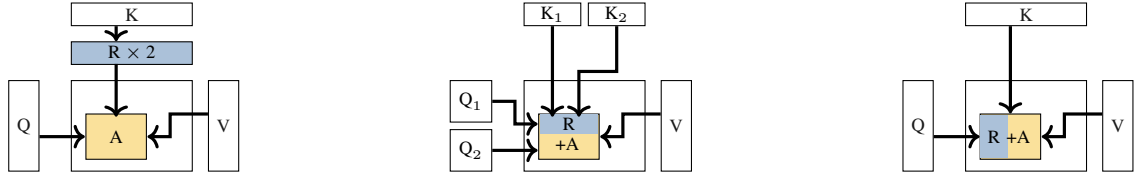

\paragraph{Naive deferred encoding is expensive} Following the idea of existing reuse approaches~\cite{ma2025blockattention,lu-etal-2025-turborag}, one straightforward implementation for deferred encoding is shown in Figure~\ref{fig:tiling}(a).
Specifically, whenever we load a KV block, we first reset its position by rotating the beginning of $\mathbf{K}$ back to position~$0$
and then rotate it forward to the target position.\footnote{Equivalently, two
successive rotary transforms per block.} Now we analyze its cost using tiling. Focus on the computation for a set of Q, K, and V tiles, where the shape of a query tile is $(M, D)$ and 
the shape of a K/V tile is $(N,D)$. We note that $M$ is the number of tokens in a Q tile, $N$ is the number of tokens in a K/V tile, and $D$ is the head size.
Then the main cost is from two general matrix multiplications (GEMMs), i.e., $\rmQ\rmK^\top$ and $\rmP\rmV$ where $\rmP=\operatorname{Softmax}(\frac{{\rmQ\rmK^\top}}{\sqrt{d_k}})$ and $d_k$ is the dimension of key ($d_k = D$ here). We always perform these two GEMMs with $4MND$ floating point operations (FLOPs).\footnote{We count multiply and add as two FLOPs.} Applying RoPE requires a 2D rotation for every pair of feature dimensions, incurring 6 FLOPs per pair (or 3 FLOPs per scalar), derived from the 4 multiplications and 2 additions in the rotation matrix application. Therefore, 
\emph{two} rotations on a K tile would introduce $\Delta\text{FLOPs}_\text{naive} \;=\; 2 \times 3ND = 6DN$ FLOPs. We have the relative extra cost is,
\[
\frac{\Delta\text{FLOPs}_\text{naive}}{\text{FLOPs}_\text{baseline}}
\;\approx\;
\frac{6DN}{4MND} \;=\; \frac{3}{2M}.
\]

For \textit{decoding}, where $M=1$ as LLMs generate token one by one, then the deferred rotation introduces {150\%} extra FLOPs. For \textit{prefilling} with $M=128$ (default in vLLM when prefilling), the extra FLOPs are still 1.17\%.
On the I/O side, each rotation needs a $\cos/\sin$ vector of length $D$ per token, hence
two rotations add $2ND$ elements. Relative to reading three tiles, the extra bandwidth fraction is
\[
\frac{\Delta\text{I/O}_\text{naive}}{\text{I/O}_\text{baseline}}
\;\approx\;
\frac{2ND}{MD + 2ND} \;=\; \frac{2N}{M+2N}.
\]
For \textit{decoding} with $N=16$ (default in vLLM when decoding), the extra IO would be approaching {100\%}, which is unaffordable. We note that the overhead of loading a Q tile is usually amortized since it can be used for multiple K tiles. 

\paragraph{Why this gets amplified on GPUs}
Modern attention kernels are often optimized to operate close to hardware limits, making them highly sensitive to modifications in the inner loop. Additional work, especially uncoalesced memory accesses, can introduce non-negligible overhead. Therefore, even seemingly minor changes, such as adding scalar loads to the decoding critical path, may noticeably degrade latency due to pipeline stalls and reduced warp occupancy.

Given the high overhead incurred by a naive implementation, we must find an efficient way to implement deferred positional encoding with low overhead.

\subsection{Efficient Rotation with Tiling}
\label{sec:rotation-within-tiling}

We implement \ourmethod\ within attention kernels via careful system-algorithm co-design to reduce the overhead.
First, we review a key property of RoPE.

\begin{fact}[Relative rotation for RoPE]
\label{fact:rope-rel}
RoPE can be represented as a rotation matrix $\rmR$ applied to queries and keys,
with the rotation angle determined by token positions. For any token pair from
$\rmQ$ and $\rmK$ with positions $\rvm$ and $\rvn$,
\[
(\rmR_{\rvm}\rvq)^\top (\rmR_{\rvn}\rvk)
= \rvq^\top \rmR_{\rvm}^\top \rmR_{\rvn}\,\rvk
= \rvq^\top \rmR_{\rvn-\rvm}\,\rvk,
\]
i.e., attention depends only on the \emph{relative} offset $\rvn-\rvm$.
\end{fact}

\paragraph{Rotate $\rmQ$ or $\rmK$} Since only relative offset matters, to change the relative position between $\rmQ$ and $\rmK$, we can rotate either $\rmQ$ or $\rmK$. We note that rotate both of them is unnecessary and increases overhead.
The choice of which nucleus to spin depends largely on the pattern of attention computation kernels.

\paragraph{Prefilling (compute-bound)}
Due to the design of PagedAttention~\cite{vllm}, prefilling kernels typically have a relative long Q tile and a short K tile, e.g., $M=128$ and $N=16$, hence \emph{rotating $\rmK$} is
cheaper than rotating $\rmQ$. We apply a \emph{single} relative rotation with
offset $\Delta$ (never ``back-to-zero then forward'') like Figure~\ref{fig:tiling}(b), combining the two half
dimensions as
\[
k_1' = k_1 \cos\Delta - k_2 \sin\Delta,\qquad
k_2' = k_1 \sin\Delta + k_2 \cos\Delta.
\]
For computation, this introduces $3$ FLOPs per scalar of a K tile. Compared with the baseline $\text{FLOPs}_{\text{baseline}}=4MND$, the relative overhead factor of prefilling with deferred encoding is
\begin{equation}\label{eq:prefilling}
{\epsilon_{\text{prefilling}} \;=\; \tfrac{3ND}{4MND} \;=\; \tfrac{3}{4M}}.
\end{equation}
When $M=128$, $\epsilon_{\text{prefilling}} = 0.59\%$. 
To save bandwidth, we keep a local position for each document's KV cache, i.e., indexing from position $0$. Therefore, when reusing a document, a fixed offset can be used for the whole document. Then we table-drive $\cos/\sin$ and load a
$D$-length row once per document, adding small additional bandwidth $\approx \frac{D}{2DN}=\frac{1}{2N}$. As shown in Figure~\ref{fig:tiling}(b), we further divide each tile into two pieces for convenient rotation. Here, we omit the overhead of loading a Q tile. The overhead ratio for bandwidth will not exceed 3.1\%. We note that since the iteration of K tiles is in the inner loop~\cite{flashattention2}, rotating $\rmK$ would not introduce a high register pressure, which can be crucial for compute-bound kernels. 

\paragraph{Decoding (bandwidth-bound)} For decoding, the attention computation kernel has a different mode because the length of a q tile is strictly $1$. While rotating a K tile would cost $3ND$ FLOPs, rotating a Q tile would cost a fixed $3D$ FLOPs. Besides, we \emph{rotate $\rmQ$} only when encountering different documents as shown in Figure~\ref{fig:tiling}(c). Let $r$ be the
fraction of KV tiles whose relative offset changes, and $B$ be the number of blocks in a document, $r=\frac{1}{B}$. Rotating a singe Q tile costs
$3MD$, giving per-trigger overhead $\tfrac{3MD}{4MND} = \frac{3}{4N}$. When $N=64$, the relative overhead factor is $1.17\%$. Averaged across all tiles, this becomes

\begin{equation}\label{eq:decode}
    {\epsilon_{\text{decode}} \;=\; r \cdot \frac{3}{4N}} = \frac{3}{4BN}.
\end{equation}

This overhead becomes negligible for large $B$; for example, with documents exceeding 1,600 tokens, $r \le 1\%$ and $\epsilon_{\text{decode}} \le 0.01\%$. To eliminate inner-loop memory traffic, we bit-pack \texttt{(block\_id, offset, mask)} into a single 64-bit register, unpacking metadata via register shifts to bypass global loads. For severe I/O intensive scenarios, we can further eliminate memory accesses by computing $\cos/\sin$ rotary vectors on-the-fly. Finally, we avoid partitioning the Q tile like the prefilling kernel shown in Figure~\ref{fig:tiling}(b), as splitting such a small tile would significantly degrade GEMM efficiency.



\subsection{Analysis: Complexity of \textsc{LazyAttention}}
\label{sec:lazy-complexity}

\paragraph{Compute}
Regarding computational complexity, \ourmethod\ preserves the asymptotic bounds of standard attention. Given sequence length $L$ and head dimension $D$, a single head performs $\mathcal{O}(L^2 D)$ FLOPs for the two core GEMMs ($\rmQ\rmK^\top$ and $\rmP\rmV$). The linear projections, which map the hidden dimension $d_{\mathrm{model}}$ to the head dimension $D$, contribute $\mathcal{O}(L d_{\mathrm{model}} D)$ but remain unaffected by our design. Consequently, deferred rotation introduces only the constant-factor overheads derived earlier: $\tfrac{3}{4M}$ for prefilling and $\tfrac{3}{4BN}$ for decoding.


\paragraph{Memory}
Regarding memory and bandwidth, we maintain a \emph{position-agnostic} KV cache logically, storing keys and values uniquely per token with $\mathcal{O}(LD)$ cost per head. In contrast, approaches that embed absolute positions must re-materialize cache entries for each shift, inflating memory usage proportional to the number of reused offsets. Decoupling position from KV data eliminates this redundancy, preserving a footprint linear in $L$.


\subsection{Analysis: TTFT and Decoding Overhead}
\label{sec:ttft-decoding}

We analyze the overhead of \ourmethod{} on the two critical metrics of inference: Time-to-First-Token (TTFT) and Inter-Token Latency.
Let $L$ be the total sequence length (including reused documents), $L'$ be the length of the new prompt segments, $D$ be the head size, $M$ be the query tile size, and $N$ be the KV tile size.
We adopt a roofline model where a kernel's time cost is $T \approx \max(F/\mathcal{P}, B/\mathcal{B})$.

\paragraph{TTFT}
In modern serving systems (e.g., vLLM), TTFT is determined by the \emph{prefilling kernel}, which processes the prompt and generates the first token in a single pass.
For RAG workloads, we only compute attention between the new prompt ($L'$) and the full context ($L$), resulting in $4LL'D$ FLOPs.
Unlike the naive deferred approach which costs $6DN$ FLOPs (rotating $\mathbf{K}$ twice), \ourmethod{} performs a \emph{single} relative rotation on $\mathbf{K}$ (since $N < M$ in prefilling).
This incurs only $3ND$ extra FLOPs per tile pair.
The relative computational overhead is $\epsilon_{\text{prefilling}} = \tfrac{3}{4M}$.
Regarding I/O, loading one rotary vector adds negligible overhead ($\le \frac{1}{2N}$ relative to reading $\mathbf{K}$ and $\mathbf{V}$).
Thus, the TTFT is modeled as follows,
\begin{equation}
\label{eq:ttft}
\text{TTFT} \approx T_{\text{prefilling}} \approx \max\left( \frac{4LL' D (1 + \frac{3}{4M})}{\mathcal{P}}, \frac{LD(2+\tfrac{1}{N})}{\mathcal{B}} \right)
\end{equation}
Note that the bandwidth term is dominated by reading the cached history from HBM ($\approx 2LD$), while the compute term scales with $L \cdot L'$.
Equation~\ref{eq:ttft} demonstrates that \ourmethod{} not only minimizes overhead ($\approx 0.59\%$ for $M=128$) but also preserves the efficiency gains from document reuse (where $L' \ll L$).

\paragraph{Decoding latency}
For subsequent token generation, the system switches to the decoding kernel ($M=1$), which is typically bandwidth-bound.
\ourmethod{} fuses the rotation into the inner loop, rotating the query $\mathbf{q}$ only when the relative offset changes.
Crucially, because rotation occurs in registers, the memory traffic remains at the baseline's $2LD$ bytes (reading only $\mathbf{K}$ and $\mathbf{V}$).
Thus, the decoding latency remains at the optimal roofline as original attention computation,
\begin{equation*}
T_{\text{decode}} \approx \frac{2LD}{\mathcal{B}}.
\end{equation*}
This confirms that \ourmethod{} introduces zero I/O overhead during generation.

\blue{
\subsection{Generalization Beyond Standard RoPE}\label{sec:generalization}

\ourmethod is not tied to standard RoPE. It applies whenever positional effects can be injected into attention-score computation, rather than materialized as position-adjusted KV states. This covers the common relative-position cases used in modern LLM serving, while preserving exactness for identical cached chunks.

\paragraph{RoPE-family variants}
For RoPE-family variants, including interleaved RoPE used by Llama~\citep{llama3}, scaled/NTK variants, and YaRN~\citep{rope-scaling}, the attention score can still be written as $\rvq^\top \rmR_{\rvn-\rvm}\,\rvk$ with variant-specific scaling metadata. The kernel therefore only needs token positions and lightweight encoding parameters; the overall zero-copy structure remains unchanged.

\paragraph{GQA/MQA compatibility}
Grouped-query attention (GQA)~\citep{ainslie2023gqa} and multi-query attention (MQA) change the mapping from query heads to shared KV heads, but do not alter how each $\rvq$--$\rvk$ score is computed. Thus, \ourmethod requires no algorithmic change for GQA/MQA models.

\paragraph{Score-space positional methods (ALiBi)}
The same lazy principle extends to score-space relative-position methods such as ALiBi~\citep{alibi2022}, where the score is modified by a position-dependent bias rather than a rotation. 
Linear attention is outside our current scope because it changes the attention state representation rather than only the positional score computation. With a case study over Falcon-7B~\citep{falcon2023}, a model supports both RoPE and ALiBi, we validated that the extra decoding cost is less than 0.06\%. This demonstrates the generalizability of \ourmethod accross different relative positional encoding methods.
}

\section{Evaluation}
\label{sec:experiment}

In this section, we evaluate \ourmethod\ along four research questions (RQs) to demonstrate its advantages, analyze its overhead, and stress-test the scope highlighted by reviewers. More experimental details and additional generalization studies can be found in \Cref{appx:exp}.
\begin{itemize}[]
  \item[\textbf{RQ1}] Does \ourmethod\ reduce time for the first token (TTFT) under different serving loads?
  \item[\textbf{RQ2}] For repeated documents across requests, does \ourmethod\ attain a high KV hit ratio with limited GPU memory for KV cache?
  \item[\textbf{RQ3}] What is the latency overhead of the extra deferred rotation operation of \ourmethod\ in prefilling and decoding respectively?
  \item[\textbf{RQ4}] Does \ourmethod\ preserve generation quality for different benchmark datasets?
\end{itemize}


\paragraph{Implementation} We implement \ourmethod on the top of vLLM~\citep{vLLMv1} within 5K lines in Python based on PyTorch v2.7 and CUDA 12.4. Our implementation is designed to be compatible and non-intrusive with the existing framework, allowing for easy integration into various models. Specifically, we use the vLLM v0.8.5.post1 V1 for efficient inference and model management, using its capabilities to optimize memory usage and computational efficiency. The code is available at \codebaseurl.

\paragraph{Models, Hardware, and Datasets} We evaluate \ourmethod using Tulu3-Block-FT\footnote{\url{https://huggingface.co/ldsjmdy/Tulu3-Block-FT}} which is fine-tuned for Block-Attention~\citep{ma2025blockattention} from Llama-3.1-Tulu-3-8B-SFT on a machine with 120GB RAM, an NVIDIA H100 96GB GPU (GH200 chipset)~\citep{h100}. {To demonstrate generalization, we also extend our evaluation to Llama-3.1-70B-Instruct and Qwen3-8B, and test on NVIDIA A100 and A40 GPUs (details in Appendix~\ref{appx:additional_exp}).} Our evaluation uses four QA benchmarks: 
(1) \textit{2WikiMQA}~\cite{xanh2020_2wikimultihop}, which requires reading multiple paragraphs, each treated as a document in \ourmethod; 
(2) \textit{HotpotQA}~\citep{yang-etal-2018-hotpotqa}, a multi-hop dataset requiring reasoning across supporting documents; 
(3) \textit{TriviaQA}~\citep{DBLP:conf/acl/JoshiCWZ17}, a reading comprehension benchmark with long web-page contexts; and 
(4) \textit{NarrativeQA}~\citep{kocisky-etal-2018-narrativeqa}, where questions demand understanding long narratives such as novels and scripts.

\paragraph{Baselines} 
We compare \ourmethod against the following baselines: 
(1) \textit{Prompt Cache}~\citep{DBLP:conf/mlsys/GimCLSK024}, the standard RAG model using a fixed-length cached prefix; 
(2) \textit{CacheBlend}~\citep{DBLP:conf/eurosys/cacheblend}, a masked variant of RAG that improves accuracy; 
(3) \textit{Block-Attention (vLLM)}~\citep{ma2025blockattention}, a block-based mechanism for cache efficiency, re-implemented in vLLM for fairness; 
(4) \textit{Prefix Caching}~\citep{DBLP:conf/sosp/KwonLZ0ZY0ZS23,vLLMv1}, the standard prefix caching in vLLM;
and (5) a \textit{MEPIC-like}~\cite{mepic} reimplementation that stores fully NoPE KV and applies rotary offsets inside a fused attention operator (see Appendix~\ref{appx:mepic} for details).

\pgfplotsset{
  every axis legend/.append style={
    at={(1.2,1.05)},
    anchor=south,
    legend columns=8,
    font=\tiny,
    draw=none
  },
  legend image post style={
    xscale=0.7,
    yscale=0.7
  }
}
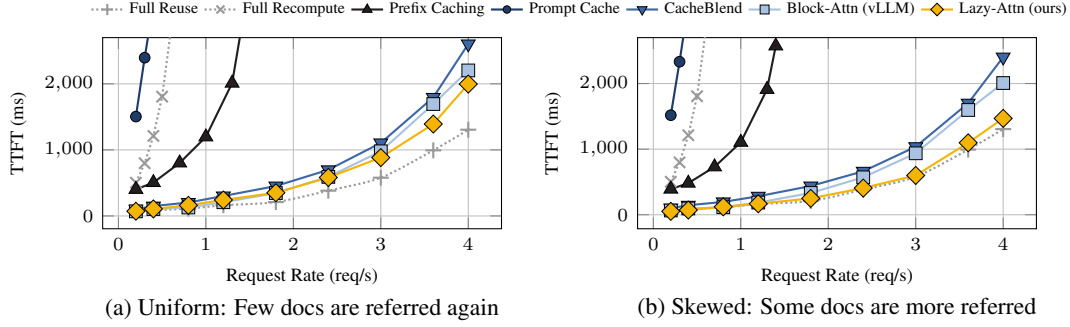
\begin{figure*}[t]
\centering
\begin{tikzpicture}
\begin{groupplot}[
  group style={
    group size=2 by 1,
    horizontal sep=1.8cm,
  },
  width=0.40\textwidth,
  height=0.24\textwidth,
  xlabel style={font=\scriptsize},
  ylabel style={font=\scriptsize},
  ticklabel style={font=\scriptsize},
  grid=both,
]

\nextgroupplot[title={\small (a) Uniform: Few docs are referred again},title style={at={(0.5,-0.4)}, anchor=north},xlabel={Request Rate (req/s)}, ylabel={TTFT (ms)},ymax=2700]

\addplot+[smooth, mark=+, thick, densely dotted, color=MyGray, mark options={solid, scale=1.5}]
  coordinates { (0.2, 71.3) (0.4, 88.1) (0.8, 110.6) (1.2, 161.4) (1.8, 209.8) (2.4, 382.5) (3.0, 578.9) (3.6, 993.2) (4.0, 1305.7) };
\addlegendentry{Full Reuse}
\addplot+[mark=x, thick, densely dotted, color=MyGray, mark options={solid, scale=1.5}]
  coordinates { (0.2, 503.2) (0.3, 798.5) (0.4, 1211.9) (0.5, 1806.3) (0.6, 2814.7) };
\addlegendentry{Full Recompute}
\addplot+[mark=triangle*, thick, color=colorset4, mark options={fill=colorset4, scale=1.5,draw=black, line width=0.3pt}]
  coordinates { (0.2, 401.7) (0.4, 505.3) (0.7, 802.4) (1.0, 1196.8) (1.3, 2010.5) (1.4, 2795.1) };
\addlegendentry{Prefix Caching}
\addplot+[mark=*, thick, solid, color=colorset3, mark options={solid, fill=colorset3, scale=1,draw=black,line width=0.3pt}]
  coordinates { (0.2, 1504.2) (0.3, 2395.8) (0.4, 3207.6) };
\addlegendentry{Prompt Cache}
\addplot+[mark=triangle*, thick, color=colorset2, mark options={fill=colorset2, scale=1.5,draw=black, line width=0.3pt,rotate=180}]
  coordinates { (0.2, 101.2) (0.4, 148.7) (0.8, 203.4) (1.2, 301.9) (1.8, 452.6) (2.4, 698.1) (3.0, 1105.3) (3.6, 1794.7) (4.0, 2603.8) };
\addlegendentry{CacheBlend}
\addplot+[mark=square*, thick, solid, color=colorset1, mark options={scale=1.2,fill=colorset1,draw=black, line width=0.3pt,},]
  coordinates { (0.2, 69.1) (0.4, 104.3) (0.8, 127.8) (1.2, 207.2) (1.8, 346.5) (2.4, 591.3) (3.0, 978.6) (3.6, 1694.4) (4.0, 2203.1) };
\addlegendentry{Block-Attn (vLLM)}
\addplot+[mark=square*, solid, thick, color=barLazy, mark options={solid, fill=barLazy, scale=1.2, rotate=45,draw=black, line width=0.3pt}]
  coordinates { (0.2, 71.5) (0.4, 106.2) (0.8, 160.4) (1.2, 241.9) (1.8, 350.7) (2.4, 582.1) (3.0, 883.5) (3.6, 1391.8) (4.0, 1996.8) };
\addlegendentry{Lazy-Attn (ours)}

\nextgroupplot[title={\small (b) Skewed: Some docs are more referred},title style={at={(0.5,-0.6)}}, xlabel={Request Rate (req/s)}, ylabel={TTFT (ms)},ymax=2700]

\addplot+[smooth, mark=+, thick, densely dotted, color=MyGray, mark options={solid, scale=1.5}]
  coordinates { (0.2, 71.3) (0.4, 88.1) (0.8, 110.6) (1.2, 161.4) (1.8, 209.8) (2.4, 382.5) (3.0, 578.9) (3.6, 993.2) (4.0, 1305.7) };

\addplot+[mark=x, thick, densely dotted, color=MyGray, mark options={solid, scale=1.5}]
  coordinates { (0.2, 503.2) (0.3, 798.5) (0.4, 1211.9) (0.5, 1806.3) (0.6, 2814.7) };

\addplot+[mark=triangle*, thick, color=colorset4, mark options={fill=colorset4, scale=1.5,draw=black, line width=0.3pt}]
  coordinates { (0.2, 390.7) (0.4, 481.6) (0.7, 730.1) (1.0, 1104.5) (1.3, 1910.43) (1.4, 2573.2) };

\addplot+[mark=*, thick, solid, color=colorset3, mark options={solid, fill=colorset3, scale=1,draw=black,line width=0.3pt}]
  coordinates { (0.2, 1517.3) (0.3, 2332.4) (0.4, 3212.53) };

\addplot+[mark=triangle*, thick, color=colorset2, mark options={fill=colorset2, scale=1.5,draw=black, line width=0.3pt,rotate=180}]
  coordinates { (0.2, 100.1) (0.4, 144.5) (0.8, 195.8) (1.2, 281.9) (1.8, 437.7) (2.4, 662.5) (3.0, 1036.7) (3.6, 1700.4) (4.0, 2403.0) };

\addplot+[mark=square*, thick, solid, color=colorset1, mark options={scale=1.2,fill=colorset1,draw=black, line width=0.3pt,},]
  coordinates { (0.2, 65.1) (0.4, 95.1) (0.8, 114.6) (1.2, 185.7) (1.8, 331.9) (2.4, 570.3) (3.0, 936.7) (3.6, 1597.0) (4.0, 2007.9) };

\addplot+[mark=square*, solid, thick, color=barLazy, mark options={solid, fill=barLazy, scale=1.2, rotate=45,draw=black, line width=0.3pt}]
  coordinates { (0.2, 51.5) (0.4, 71.7) (0.8, 120.65) (1.2, 167.5) (1.8, 249.4) (2.4, 406.6) (3.0, 599.5) (3.6, 1096.7) (4.0, 1468.7) };

\end{groupplot}
\end{tikzpicture}

\vspace{-2mm}

\caption{{TTFT vs\ request rate under different document distributions.}
We vary the request rate (req/s) and plot time-to-first-token (TTFT).
(a) \emph{Uniform}—few documents recur across requests (low reuse).
(b) \emph{Skewed}—a small set of documents is frequently reused (high reuse).}~\label{fig:ttft-main}

\end{figure*}

\begin{table*}[t]
    \centering
    \begin{threeparttable}
    \caption{VRAM cache hit ratio (\%) under different KV cache budgets.
    A high hit ratio indicates more effective cache reuse. \ourmethod\ is the most effective,
    and its advantage persists up to the full GPU budget (a single-copy design keeps reusing
    documents at any offset, whereas the gains of position-coupled methods saturate).}
    \scriptsize
    \setlength{\tabcolsep}{0pt}
    \begin{tabular*}{\textwidth}{@{\extracolsep{\fill}}lrrrrrrrrrrrrrrr@{}}
        \toprule
        KV Cache Mem Size ($\rightarrow$) & \multicolumn{3}{c}{1 GB} & \multicolumn{3}{c}{5 GB} & \multicolumn{3}{c}{10 GB} & \multicolumn{3}{c}{50 GB} & \multicolumn{3}{c}{No-limit$^\dagger$} \\
        \cmidrule(lr){2-4} \cmidrule(lr){5-7} \cmidrule(lr){8-10} \cmidrule(lr){11-13} \cmidrule(lr){14-16}
        Document Skewness ($\rightarrow$) & Low & Mid & High & Low & Mid & High & Low & Mid & High & Low & Mid & High & Low & Mid & High\\
        \midrule
        Prefix Caching
        & 0.00 & 0.00 & 0.00
        & 0.04 & 0.30 & 0.70
        & 0.08 & 0.55 & 1.22
        & 0.46 & 1.95 & 3.07
        & 0.58 & 2.16 & 3.25 \\
        CacheBlend
        & 1.51 & 4.95 & 5.96
        & 6.29 & 14.43 & 14.95
        & 8.91 & 17.33 & 16.78
        & 14.32 & 21.87 & 19.06
        & 15.21 & 22.45 & 19.36 \\
        Block-Attention (vLLM)
        & 1.84 & 6.03 & 7.27
        & 7.67 & 17.59 & 18.23
        & 10.86 & 21.13 & 20.47
        & 17.46 & 26.67 & 23.24
        & 18.55 & 27.38 & 23.61 \\
        \textit{LazyAttention (Ours)}
        & \textbf{3.47} & \textbf{11.11} & \textbf{13.57}
        & \textbf{10.85} & \textbf{21.16} & \textbf{20.49}
        & \textbf{13.78} & \textbf{23.89} & \textbf{21.92}
        & \textbf{20.22} & \textbf{28.44} & \textbf{24.14}
        & \textbf{21.33} & \textbf{29.09} & \textbf{24.50} \\
        \bottomrule
    \end{tabular*}
    \begin{tablenotes}\footnotesize
    \item[$\dagger$] No-limit: cache bounded only by GPU memory --- NVIDIA H100 96\,GB GPU (GH200 chipset) at 0.9 util, $\approx$66\,GB usable KV for the 8B model. All cells are trace-driven simulation (Zipf doc popularity, skew $\alpha=1.1/1.5/2.1$ for Low/Mid/High; 100k-doc pool, $K=10$ docs/query).
    \end{tablenotes}
    \label{tab:hit_mem}
    \end{threeparttable}
\end{table*}

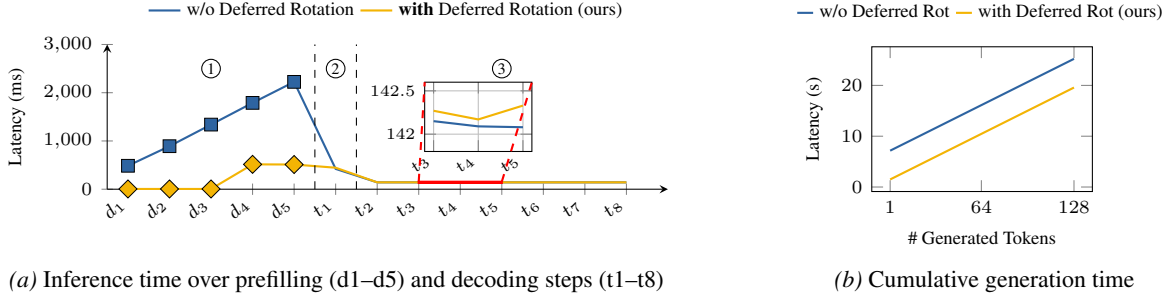
\begin{figure*}[t]
  \centering
  \hfill
  \begin{subfigure}[t]{0.62\textwidth}
  \centering
  \begin{tikzpicture}

    \begin{axis}[
        name=mainplot, 
        width=9cm,
        height=3.5cm,
        ylabel={Latency (ms)},
        ymin=0, ymax=3000,
        xmin=0.5, xmax=14,
        axis lines=left,
        legend style={
            at={(0.5,1.1)},
            anchor=south,
            legend columns=-1,
            font=\scriptsize
        },
        xtick={1,2,3,4,5, 6, 7,8,9,10,11,12,13},
        xticklabels={
          $d_1$,$d_2$,$d_3$,$d_4$,$d_5$,
          $t_1$,
          $t_2$,$t_3$,$t_4$,$t_5$,$t_6$,$t_7$,$t_8$,
        },
        xticklabel style={rotate=45,anchor=east},
        font=\scriptsize,
        tick label style={font=\scriptsize},
        grid=none,
    ]

    \addplot[mark=none, thick, solid, color=barFull] coordinates {
        (1,485.5) (2,891.6) (3,1339.3) (4,1789.3) (5,2225.0)
        (6, 421.15) (7, 142.21) (8, 142.15) (9,142.09)
        (10,142.08) (11,142.12) (12,142.04) (13,141.86)
    };
    \addlegendentry{w/o Deferred Rotation}

    \addplot[mark=none, solid, thick, color=barLazy] coordinates {
        (1,5) (2,5) (3, 5) (4, 515.0) (5, 511.9)
        (6,445.54) (7, 142.54) (8,142.27) (9,142.17)
        (10,142.33) (11,142.22) (12,142.10) (13,142.22)
    };
    \addlegendentry{\textbf{with} Deferred Rotation (ours)}

    \addplot[only marks, mark=square*, color=barFull, mark options={scale=1.2,fill=barFull,draw=black, line width=0.3pt}, forget plot] coordinates {
        (1,485.5) (2,891.6) (3,1339.3) (4,1789.3) (5,2225.0) 
    };

    \addplot[only marks, mark=square*, color=barLazy, mark options={solid, fill=barLazy, scale=1.2, rotate=45,draw=black, line width=0.3pt}, forget plot] coordinates {
        (1,5) (2,5) (3, 5) (4, 515.0) (5, 511.9) 
    };
    
    \draw[dashed] (axis cs:5.5,0) -- (axis cs:5.5,3000);
    \draw[dashed] (axis cs:6.5,0) -- (axis cs:6.5,3000);
    \node at (axis cs:3,2500) {\circled{\tiny 1}};
    \node at (axis cs:6,2500) {\circled{\tiny 2}};
    \node at (axis cs:10,2500) {\circled{\tiny 3}};

    \coordinate (zoom_rect_sw) at (axis cs:8, 130);
    \coordinate (zoom_rect_ne) at (axis cs:10, 155);

    \end{axis} 

    \begin{axis}[
        name=zoomplot, 
        at={(mainplot.north east)}, 
        anchor=north east,
        xshift=-18mm, yshift=-5mm, 
        width=3cm, 
        height=2.5cm,
        xmin=7.8, xmax=10.2,
        ymin=141.8, ymax=142.6, 
        grid=major,
        font=\tiny,
        ytick={142, 142.5},
        xtick={8,9,10},
        xticklabels={$t_3$,$t_4$,$t_5$},
        xticklabel style={rotate=45,anchor=east},
        yticklabel style={/pgf/number format/fixed, /pgf/number format/precision=2} 
    ]
        \addplot[mark=none, thick, solid, color=barFull] coordinates {
 (8, 142.15) (9,142.09)
            (10,142.08) 
        };
        \addplot[mark=none, solid, thick, color=barLazy] coordinates {
(8,142.27) (9,142.17)
            (10,142.33) 
        };
    \end{axis} 

    \draw[red, thick] (zoom_rect_sw) rectangle (zoom_rect_ne);
    \coordinate (zoom_rect_nw) at (zoom_rect_sw |- zoom_rect_ne); 
    \draw[red, thick, dashed] (zoom_rect_nw) -- (zoomplot.north west);
    \draw[red, thick, dashed] (zoom_rect_ne) -- (zoomplot.north east);

\end{tikzpicture}
    \subcaption{\small Inference time over prefilling (d1--d5) and decoding steps (t1--t8)}
  \end{subfigure}
  \hfill
  \begin{subfigure}[t]{0.35\textwidth}
  \centering
  \begin{tikzpicture}
\begin{axis}[
scaled x ticks=false,
    width=4.5cm, height=3.5cm,
    xlabel={\# Generated Tokens},
    ylabel={Latency (s)},
    xtick={1,64,128},
        font=\scriptsize,
        tick label style={font=\scriptsize},
        grid=none,
    legend style={
        at={(0.5,1.1)},
        anchor=south,
        legend columns=-1,
        font=\scriptsize
    },
    grid=none,
    font=\scriptsize,
    legend image post style={xscale=0.5, yscale=1},
]
\addplot[barFull, thick] coordinates {
(1, 7.15185) (8, 8.146434952946752) (16, 9.281862667039409) (128, 25.223472468996793)
};
\addlegendentry{w/o Deferred Rot}

\addplot[barLazy, thick] coordinates {
(1, 1.48744) (8, 2.4827249529467523) (16, 3.6189526670394097)  (128, 19.57351)
};
\addlegendentry{with Deferred Rot (ours)}

\end{axis}
\end{tikzpicture}
    \subcaption{\small Cumulative generation time}~\label{subfig:accum}
  \end{subfigure}
  \hfill
  
  \vspace{-1em}
  \caption{Overhead analysis. (a) Breakdown of overhead across phases: \ding{172} document processing, \ding{173} query prefilling, and \ding{174} decoding. Deferred rotation adds only 0.13\% overhead in decoding. (b) Accumulated latency with and without deferred rotation, where the dominant gain comes from document processing rather than token generation.}
  \label{fig:overhead}
\end{figure*}

\subsection{Higher Responsiveness with Shorter TTFT}
\label{sec:rq1-ttft}

We evaluate system responsiveness by generating request streams with controlled arrival rates (req/s), sampling from a mixture of four datasets. Each request queries five documents whose KV states have been precomputed. We assess two distinct traffic regimes: \emph{Uniform} sampling (representing low reuse potential) and \emph{Skewed} sampling (Zipfian distribution with $\alpha=2.1$, representing high reuse). All methods operate under identical batching policies, paging parameters, and memory budgets. We report mean TTFT and compare against key baselines, bounded by \emph{Full Reuse} (theoretical lower bound) and \emph{Full Recompute} (upper bound).

Under \emph{Uniform} traffic (Figure~\ref{fig:ttft-main}a), where reuse opportunities are naturally scarce, \ourmethod{} remains highly competitive. It tracks \emph{Block-Attn} closely at low-to-moderate loads and consistently outperforms \emph{Prefix Caching}, \emph{PromptCache}, and \emph{CacheBlend}. This performance validates our kernel analysis in Section~\ref{sec:ttft-decoding}: the prefilling overhead is minimal ($\tfrac{3}{4M}$) and the decoding overhead averages $r\cdot \tfrac{3}{4N}$. Since the fraction $r$ of nonzero-offset tiles is typically small and the tile size $N$ is adequate (e.g., 64), our method incurs negligible cost even when reuse is absent.

Under \emph{Skewed} traffic (Figure~\ref{fig:ttft-main}b), where a small set of documents is frequently accessed at varying positions, \ourmethod{} demonstrates superior scalability. We achieve significantly lower TTFT and sustain higher throughput before saturation. This advantage stems directly from \emph{position-agnostic reuse}: \ourmethod{} utilizes the same physical KV blocks across different logical offsets without materializing new copies. In contrast, baselines face structural inefficiencies: \emph{CacheBlend} incurs reconstruction overheads, \emph{PromptCache} expands prompt length, and \emph{Block-Attn} suffers from cache fragmentation due to storing duplicate, position-dependent KV copies. {We observe consistent speedups on the larger Llama-3.1-70B model and across different hardware configurations (Appendix~\ref{appx:largermodel}).}

\begin{table*}[t]
\caption{Question-answer accuracy for various benchmark datasets. \emph{Exact match} scores are reported.
Our \ourmethod performs the identical computations as Block-Attention, achieving nearly identical scores, where slight differences are due to tokenization and limited precision in floating-point operations.
That is, \ourmethod significantly reduces TTFT  and increases the reuse opportunities (as reported before) with negligible accuracy loss.}
\centering
{\small
    \begin{tabular}{lccccc} 
        \toprule
        Dataset & Full-Attn & CacheBlend & Block-Attn & \textit{Block-Attn (vLLM)} & \textit{\textbf{Lazy-Attn (ours)}} \\
        \midrule
        2WikiMQA
        & 73.6 & 71.1 & 72.2 & 71.4 & 70.7\\
        TriviaQA
         & 75.2 & 69.2 & 72.3 & 72.1 & 73.0  \\
        NarrativeQA
        & 62.2 & 60.1 & 60.4 & 61.0 & 59.7  \\
        HotpotQA
        & 76.2 & 69.7 & 75.1 & 72.5 & 73.3 \\ 
        \midrule
        Average & 71.8 & 67.5 & 71.2 & 69.3 & 69.2 \\
        \bottomrule
    \end{tabular}
}
    \label{tab:comparison_results_example} 
\end{table*}

\subsection{Higher Cache Hit Ratios}
\label{sec:rq2-hit}


We evaluate cache efficiency by varying both the KV cache budget (1/5/10/50\,GB and a
no-limit setting constrained only by available GPU memory) and document popularity skew
(Low/Mid/High with $\alpha=1.1/1.5/2.1$). Our primary metric is the VRAM cache hit ratio,
defined as the fraction of KV-block lookups served directly from cache without recomputation.
As shown in Table~\ref{tab:hit_mem}, \ourmethod\ achieves the highest hit ratio in every
budget--skew configuration. The advantage is most pronounced---in relative terms---under tight
memory, where avoiding position-coupled duplicates frees scarce VRAM for additional documents:
at 1\,GB, \ourmethod\ nearly doubles the strongest baseline (13.57\% vs.\ 7.27\% at high skew,
3.47\% vs.\ 1.84\% at low skew). As the budget grows, all methods improve and the gap narrows,
yet \ourmethod\ retains the lead at every skew level---e.g., 23.89\% versus 21.13\% for
Block-Attention at 10\,GB and mid skew. Crucially, the benefit does not disappear once memory is
abundant: even in the no-limit, full-GPU setting, \ourmethod\ leads at all skews (29.09\% vs.\
27.38\% at mid skew, 21.33\% vs.\ 18.55\% at low skew). These results confirm that storing a
single physical KV copy per document, reusable at arbitrary offsets, makes position-agnostic
reuse valuable across the entire memory-budget spectrum---most decisively when VRAM is the
binding constraint, and still beneficial when it is not.


\subsection{Fused kernel efficiency: our rotation overhead is negligible}
\label{sec:overhead}

To isolate kernel-level costs, we construct a single long RAG request with
five documents (each 4{,}096 tokens) and a 64-token query.
The serving system uses a 2{,}048-token chunk budget per pass.
We preload three documents' KV blocks in DRAM to emulate ``hot'' content and
leave the remaining documents ``cold.'' Common preamble is omitted. We compare two ablations:
\emph{w/o Deferred Rotation} (no position-agnostic reuse; blocks are processed
in the conventional pipeline) and \emph{with Deferred Rotation (ours)}, which
reuses cached KV at arbitrary offsets and performs the necessary RoPE rotation
inside the fused attention kernel. Scheduler, batching, paging, and launch
parameters are identical; only the rotation path differs.

As shown in Figure~\ref{fig:overhead}, we analyze the overhead across three key phases. {\ding{172} Document processing:} our method reduces latency to near-zero for hot documents by reusing KV cache without duplication, whereas recomputation costs dominate the baseline. 
{\ding{173} Query prefilling:} the performance remains comparable to the baseline; our kernel rotates keys just once per KV tile using the relative offset, introducing only a marginal $\tfrac{3}{4M}$ increase in FLOPs (Section~\ref{sec:rotation-within-tiling}). 
{\ding{174} Decoding:} the latency curves overlap, with the zoomed inset revealing a negligible per-token overhead of \textit{0.13\%}. This aligns with our theoretical analysis of $r\cdot\frac{3}{4N}$ (Section~\ref{sec:ttft-decoding}), given the small fraction $r$ of nonzero-offset tiles. 
Figure~\ref{fig:overhead}~(b) confirms that the cumulative generation gap remains constant as the token count increases, proving that decoding overhead does not accumulate. We further verify that this efficiency holds for document lengths up to 16K and context lengths up to 128K (Appendix~\ref{appendix:scaling-length}).

\subsection{{Generation Quality}}

We evaluate \ourmethod{} following the setting of Block-Attention~\citep{ma2025blockattention}, assessing QA generation quality via Exact Match (EM). Unless otherwise specified, we use the same model checkpoint, prompts, and decoding hyperparameters. Empirically, \ourmethod{} attains EM scores comparable to Block-Attention, confirming that deferred positional encoding preserves model fidelity. Because we serve all methods on the same vLLM engine, the absolute EM scores sit slightly below those reported by \citet{ma2025blockattention} under HuggingFace \texttt{transformers}; this offset arises from differences in numerical precision and attention-kernel implementation in vLLM rather than from any particular method, and it shifts all methods equally. We further validate \ourmethod{} on a long-form literature-review task and a few-shot classification workload, demonstrating that it accelerates recurring-chunk workloads beyond simple QA (Appendices~\ref{appx:beyond_rag} and~\ref{appx:long_context_constrained}).

\section{Conclusion}\label{sec:conclusion}

In this paper, we proposed \ourmethod, a novel deferred positional encoding mechanism for Transformer models. Our approach decouples positional encoding from the KV cache, allowing for more efficient caching and improved performance in retrieval-augmented generation tasks. We demonstrated the effectiveness of our method through extensive experiments on various datasets, showing significant improvements in TTFT, cache hit ratio, and overall model performance. Our findings suggest that \ourmethod is a practical solution for enhancing the efficiency and accuracy of Transformer models in long-context tasks across different kinds of relative positional encoding methods.

\section*{Impact Statement}
This paper presents work aimed at improving the efficiency of LLM inference, specifically for RAG. By enabling position-agnostic KV reuse, our method significantly reduces the computational overhead and memory bandwidth required for processing long contexts. This contributes to LLM serving by reducing the energy consumption associated with serving large-scale RAG applications. While our work lowers the barrier to deploying LLMs, potentially facilitating both beneficial and harmful applications, it does not introduce new safety risks beyond those inherent to the underlying models.




\section*{Acknowledgements}

This work was supported in part by the National Science Foundation under grants \#2312561 and \#2440498. This work used Delta and DeltaAI at the National Center for Supercomputing Applications (NCSA) through allocation CIS240661 from the Advanced Cyberinfrastructure Coordination Ecosystem: Services \& Support (ACCESS) program, which is supported by U.S.\ National Science Foundation grants \#2138259, \#2138286, \#2138307, \#2137603, and \#2138296. This work was also supported by the IBM-Illinois Discovery Accelerator Institute.
\balance
\bibliography{main}

@inproceedings{fang2025large,
  title={Large-scale Evaluation of Notebook Checkpointing with AI Agents},
  author={Fang, Hanxi and Chockchowwat, Supawit and Sundaram, Hari and Park, Yongjoo},
  booktitle={Proceedings of the Extended Abstracts of the CHI Conference on Human Factors in Computing Systems},
  pages={1--8},
  year={2025}
}

@article{pbemeets,
  title={PBE Meets LLM: When Few Examples Aren't Few-Shot Enough},
  author={Zhang, Shuning and Park, Yongjoo},
  journal={arXiv preprint arXiv:2507.05403},
  year={2025}
}

@inproceedings{
hu2025epic,
title={{EPIC}: Efficient Position-Independent Caching for Serving Large Language Models},
author={Junhao Hu and Wenrui Huang and Weidong Wang and Haoyi Wang and tiancheng hu and zhang qin and Hao Feng and Xusheng Chen and Yizhou Shan and Tao Xie},
booktitle={Forty-second International Conference on Machine Learning},
year={2025},
url={https://openreview.net/forum?id=qjd3ZUiHRT}
}

@misc{yang2025qwen3technicalreport,
      title={Qwen3 Technical Report}, 
      author={An Yang and Anfeng Li and Baosong Yang and Beichen Zhang and Binyuan Hui and Bo Zheng and Bowen Yu and Chang Gao and Chengen Huang and Chenxu Lv and Chujie Zheng and Dayiheng Liu and Fan Zhou and Fei Huang and Feng Hu and Hao Ge and Haoran Wei and Huan Lin and Jialong Tang and Jian Yang and Jianhong Tu and Jianwei Zhang and Jianxin Yang and Jiaxi Yang and Jing Zhou and Jingren Zhou and Junyang Lin and Kai Dang and Keqin Bao and Kexin Yang and Le Yu and Lianghao Deng and Mei Li and Mingfeng Xue and Mingze Li and Pei Zhang and Peng Wang and Qin Zhu and Rui Men and Ruize Gao and Shixuan Liu and Shuang Luo and Tianhao Li and Tianyi Tang and Wenbiao Yin and Xingzhang Ren and Xinyu Wang and Xinyu Zhang and Xuancheng Ren and Yang Fan and Yang Su and Yichang Zhang and Yinger Zhang and Yu Wan and Yuqiong Liu and Zekun Wang and Zeyu Cui and Zhenru Zhang and Zhipeng Zhou and Zihan Qiu},
      year={2025},
      eprint={2505.09388},
      archivePrefix={arXiv},
      primaryClass={cs.CL},
      url={https://arxiv.org/abs/2505.09388}, 
}

@inproceedings{lu-etal-2025-turborag,
    title = "{T}urbo{RAG}: Accelerating Retrieval-Augmented Generation with Precomputed {KV} Caches for Chunked Text",
    author = "Lu, Songshuo  and
      Wang, Hua  and
      Rong, Yutian  and
      Chen, Zhi  and
      Tang, Yaohua",
    editor = "Christodoulopoulos, Christos  and
      Chakraborty, Tanmoy  and
      Rose, Carolyn  and
      Peng, Violet",
    booktitle = "Proceedings of the 2025 Conference on Empirical Methods in Natural Language Processing",
    month = nov,
    year = "2025",
    address = "Suzhou, China",
    publisher = "Association for Computational Linguistics",
    url = "https://aclanthology.org/2025.emnlp-main.334/",
    doi = "10.18653/v1/2025.emnlp-main.334",
    pages = "6599--6612",
    ISBN = "979-8-89176-332-6"
}

@inproceedings{ainslie2023gqa,
  title={GQA: Training Generalized Multi-Query Transformer Models from Multi-Head Checkpoints},
  author={Ainslie, Joshua and Lee-Thorp, James and de Jong, Michiel and Zemlyanskiy, Yury and Lebron, Federico and Sanghai, Sumit},
  booktitle={Proceedings of the 2023 Conference on Empirical Methods in Natural Language Processing},
  pages={4895--4901},
  year={2023}
}

@article{minicache,
  title={Minicache: Kv cache compression in depth dimension for large language models},
  author={Liu, Akide and Liu, Jing and Pan, Zizheng and He, Yefei and Haffari, Reza and Zhuang, Bohan},
  journal={Advances in Neural Information Processing Systems},
  volume={37},
  pages={139997--140031},
  year={2024}
}

@inproceedings{
kvcompress1,
title={Model Tells You What to Discard: Adaptive {KV} Cache Compression for {LLM}s},
author={Suyu Ge and Yunan Zhang and Liyuan Liu and Minjia Zhang and Jiawei Han and Jianfeng Gao},
booktitle={The Twelfth International Conference on Learning Representations},
year={2024},
url={https://openreview.net/forum?id=uNrFpDPMyo}
}

@ARTICLE{gemini,
       author = {{Gemini Team} and {Georgiev}, Petko and {Lei}, Ving Ian and {Burnell}, Ryan and {Bai}, Libin and {Gulati}, Anmol and {Tanzer}, Garrett and {Vincent}, Damien and {Pan}, Zhufeng and {Wang}, Shibo and {Mariooryad}, Soroosh and {Ding}, Yifan and {Geng}, Xinyang and {Alcober}, Fred and {Frostig}, Roy and {Omernick}, Mark and {Walker}, Lexi and {Paduraru}, Cosmin and {Sorokin}, Christina and {Tacchetti}, Andrea and {Gaffney}, Colin and {Daruki}, Samira and {Sercinoglu}, Olcan and {Gleicher}, Zach and {Love}, Juliette and {Voigtlaender}, Paul and {Jain}, Rohan and {Surita}, Gabriela and {Mohamed}, Kareem and {Blevins}, Rory and {Ahn}, Junwhan and {Zhu}, Tao and {Kawintiranon}, Kornraphop and {Firat}, Orhan and {Gu}, Yiming and {Zhang}, Yujing and {Rahtz}, Matthew and {Faruqui}, Manaal and {Clay}, Natalie and {Gilmer}, Justin and {Co-Reyes}, JD and {Penchev}, Ivo and {Zhu}, Rui and {Morioka}, Nobuyuki and {Hui}, Kevin and {Haridasan}, Krishna and {Campos}, Victor and {Mahdieh}, Mahdis and {Guo}, Mandy and {Hassan}, Samer and {Kilgour}, Kevin and {Vezer}, Arpi and {Cheng}, Heng-Tze and {de Liedekerke}, Raoul and {Goyal}, Siddharth and {Barham}, Paul and {Strouse}, DJ and {Noury}, Seb and {Adler}, Jonas and {Sundararajan}, Mukund and {Vikram}, Sharad and {Lepikhin}, Dmitry and {Paganini}, Michela and {Garcia}, Xavier and {Yang}, Fan and {Valter}, Dasha and {Trebacz}, Maja and {Vodrahalli}, Kiran and {Asawaroengchai}, Chulayuth and {Ring}, Roman and {Kalb}, Norbert and {Baldini Soares}, Livio and {Brahma}, Siddhartha and {Steiner}, David and {Yu}, Tianhe and {Mentzer}, Fabian and {He}, Antoine and {Gonzalez}, Lucas and {Xu}, Bibo and {Lopez Kaufman}, Raphael and {El Shafey}, Laurent and {Oh}, Junhyuk and {Hennigan}, Tom and {van den Driessche}, George and {Odoom}, Seth and {Lucic}, Mario and {Roelofs}, Becca and {Lall}, Sid and {Marathe}, Amit and {Chan}, Betty and {Ontanon}, Santiago and {He}, Luheng and {Teplyashin}, Denis and {Lai}, Jonathan and {Crone}, Phil and {Damoc}, Bogdan and {Ho}, Lewis and {Riedel}, Sebastian and {Lenc}, Karel and {Yeh}, Chih-Kuan and {Chowdhery}, Aakanksha and {Xu}, Yang and {Kazemi}, Mehran and {Amid}, Ehsan and {Petrushkina}, Anastasia and {Swersky}, Kevin and {Khodaei}, Ali and {Chen}, Gowoon and {Larkin}, Chris and {Pinto}, Mario and {Yan}, Geng and {Puigdomenech Badia}, Adria and {Patil}, Piyush and {Hansen}, Steven and {Orr}, Dave and {Arnold}, Sebastien M.~R. and {Grimstad}, Jordan and {Dai}, Andrew and {Douglas}, Sholto and {Sinha}, Rishika and {Yadav}, Vikas and {Chen}, Xi and {Gribovskaya}, Elena and {Austin}, Jacob and {Zhao}, Jeffrey and {Patel}, Kaushal and {Komarek}, Paul and {Austin}, Sophia and {Borgeaud}, Sebastian and {Friso}, Linda and {Goyal}, Abhimanyu and {Caine}, Ben and {Cao}, Kris and {Chung}, Da-Woon and {Lamm}, Matthew and {Barth-Maron}, Gabe and {Kagohara}, Thais and {Olszewska}, Kate and {Chen}, Mia and {Shivakumar}, Kaushik and {Agarwal}, Rishabh and {Godhia}, Harshal and {Rajwar}, Ravi and {Snaider}, Javier and {Dotiwalla}, Xerxes and {Liu}, Yuan and {Barua}, Aditya and {Ungureanu}, Victor and {Zhang}, Yuan and {Batsaikhan}, Bat-Orgil and {Wirth}, Mateo and {Qin}, James and {Danihelka}, Ivo and {Doshi}, Tulsee and {Chadwick}, Martin and {Chen}, Jilin and {Jain}, Sanil and {Le}, Quoc and {Kar}, Arjun and {Gurumurthy}, Madhu and {Li}, Cheng and {Sang}, Ruoxin and {Liu}, Fangyu and {Lamprou}, Lampros and {Munoz}, Rich and {Lintz}, Nathan and {Mehta}, Harsh and {Howard}, Heidi and {Reynolds}, Malcolm and {Aroyo}, Lora and {Wang}, Quan and {Blanco}, Lorenzo and {Cassirer}, Albin and {Griffith}, Jordan and {Das}, Dipanjan and {Lee}, Stephan and {Sygnowski}, Jakub and {Fisher}, Zach and {Besley}, James and {Powell}, Richard and {Ahmed}, Zafarali and {Paulus}, Dominik and {Reitter}, David and {Borsos}, Zalan and {Joshi}, Rishabh and {Pope}, Aedan and {Hand}, Steven and {Selo}, Vittorio and {Jain}, Vihan and {Sethi}, Nikhil and {Goel}, Megha and {Makino}, Takaki and {May}, Rhys and {Yang}, Zhen and {Schalkwyk}, Johan and {Butterfield}, Christina and {Hauth}, Anja and {Goldin}, Alex and {Hawkins}, Will and {Senter}, Evan},
        title = "{Gemini 1.5: Unlocking multimodal understanding across millions of tokens of context}",
      journal = {arXiv e-prints},
         year = 2024,
        month = mar,
          eid = {arXiv:2403.05530},
        pages = {arXiv:2403.05530},
          doi = {10.48550/arXiv.2403.05530},
archivePrefix = {arXiv},
       eprint = {2403.05530},
 primaryClass = {cs.CL},
       adsurl = {https://ui.adsabs.harvard.edu/abs/2024arXiv240305530G}
}

@inproceedings{bert,
  title={Bert: Pre-training of deep bidirectional transformers for language understanding},
  author={Devlin, Jacob and Chang, Ming-Wei and Lee, Kenton and Toutanova, Kristina},
  booktitle={Proceedings of the 2019 conference of the North American chapter of the association for computational linguistics: human language technologies, volume 1 (long and short papers)},
  pages={4171--4186},
  year={2019}
}

@inproceedings{orca,
  title={Orca: A distributed serving system for $\{$Transformer-Based$\}$ generative models},
  author={Yu, Gyeong-In and Jeong, Joo Seong and Kim, Geon-Woo and Kim, Soojeong and Chun, Byung-Gon},
  booktitle={16th USENIX Symposium on Operating Systems Design and Implementation (OSDI 22)},
  pages={521--538},
  year={2022}
}

@inproceedings{DBLP:conf/mlsys/GimCLSK024,
  author       = {In Gim and
                  Guojun Chen and
                  Seung{-}Seob Lee and
                  Nikhil Sarda and
                  Anurag Khandelwal and
                  Lin Zhong},
  editor       = {Phillip B. Gibbons and
                  Gennady Pekhimenko and
                  Christopher De Sa},
  title        = {Prompt Cache: Modular Attention Reuse for Low-Latency Inference},
  booktitle    = {Proceedings of the Seventh Annual Conference on Machine Learning and
                  Systems, MLSys 2024, Santa Clara, CA, USA, May 13-16, 2024},
  publisher    = {mlsys.org},
  year         = {2024},
  url          = {https://openreview.net/forum?id=pfNNZnJHNl},
  bibsource    = {dblp computer science bibliography, https://dblp.org}
}

@inproceedings{DBLP:conf/eurosys/cacheblend,
  author       = {Jiayi Yao and
                  Hanchen Li and
                  Yuhan Liu and
                  Siddhant Ray and
                  Yihua Cheng and
                  Qizheng Zhang and
                  Kuntai Du and
                  Shan Lu and
                  Junchen Jiang},
  title        = {CacheBlend: Fast Large Language Model Serving for RAG with Cached Knowledge Fusion},
  booktitle    = {Proceedings of the Nineteenth European Conference on Computer Systems,
                  EuroSys 2025, Rotterdam,
March 30 - April 3, 2025},
  publisher    = {{ACM}},
  year         = {2025},
  bibsource    = {dblp computer science bibliography, https://dblp.org}
}

@article{DBLP:journals/corr/abs-2404-12457,
  author       = {Chao Jin and
                  Zili Zhang and
                  Xuanlin Jiang and
                  Fangyue Liu and
                  Xin Liu and
                  Xuanzhe Liu and
                  Xin Jin},
  title        = {RAGCache: Efficient Knowledge Caching for Retrieval-Augmented Generation},
  journal      = {CoRR},
  volume       = {abs/2404.12457},
  year         = {2024},
  url          = {https://doi.org/10.48550/arXiv.2404.12457},
  doi          = {10.48550/ARXIV.2404.12457},
  eprinttype    = {arXiv},
  eprint       = {2404.12457},
  bibsource    = {dblp computer science bibliography, https://dblp.org}
}

@inproceedings{
ma2025blockattention,
title={Block-Attention for Efficient Prefilling},
author={Dongyang Ma and Yan Wang and Tian Lan},
booktitle={The Thirteenth International Conference on Learning Representations},
year={2025},
url={https://openreview.net/forum?id=7zNYY1E2fq}
}

@misc{LangChain2024,
  title        = {LangChain},
  howpublished = {\url{https://python.langchain.com/docs/get_started/introduction}},
  year         = {2024},
  note         = {Accessed: 2024-12-09}
}

@inproceedings{Lewis2020RAG,
  author    = {Patrick Lewis and Ethan Perez and Aleksandra Piktus and Fabio Petroni and Vladimir Karpukhin and Naman Goyal and Heinrich K{\"u}ttler and Mike Lewis and Wen-tau Yih and Tim Rockt{\"a}schel and others},
  title     = {Retrieval-Augmented Generation for Knowledge-Intensive NLP Tasks},
  booktitle = {Advances in Neural Information Processing Systems},
  year      = {2020}
}

@inproceedings {DBLP:journals/corr/abs-2407-00079,
author = {Ruoyu Qin and Zheming Li and Weiran He and Jialei Cui and Feng Ren and Mingxing Zhang and Yongwei Wu and Weimin Zheng and Xinran Xu},
title = {Mooncake: Trading More Storage for Less Computation {\textemdash} A {KVCache-centric} Architecture for Serving {LLM} Chatbot},
booktitle = {23rd USENIX Conference on File and Storage Technologies (FAST 25)},
year = {2025},
isbn = {978-1-939133-45-8},
address = {Santa Clara, CA},
pages = {155--170},
url = {https://www.usenix.org/conference/fast25/presentation/qin},
publisher = {USENIX Association},
month = feb
}

@inproceedings{DBLP:conf/sosp/KwonLZ0ZY0ZS23,
  author       = {Woosuk Kwon and
                  Zhuohan Li and
                  Siyuan Zhuang and
                  Ying Sheng and
                  Lianmin Zheng and
                  Cody Hao Yu and
                  Joseph Gonzalez and
                  Hao Zhang and
                  Ion Stoica},
  editor       = {Jason Flinn and
                  Margo I. Seltzer and
                  Peter Druschel and
                  Antoine Kaufmann and
                  Jonathan Mace},
  title        = {Efficient Memory Management for Large Language Model Serving with
                  PagedAttention},
  booktitle    = {Proceedings of the 29th Symposium on Operating Systems Principles,
                  {SOSP} 2023, Koblenz, Germany, October 23-26, 2023},
  pages        = {611--626},
  publisher    = {{ACM}},
  year         = {2023},
  url          = {https://doi.org/10.1145/3600006.3613165},
  doi          = {10.1145/3600006.3613165},
  bibsource    = {dblp computer science bibliography, https://dblp.org}
}

@misc{llama3,
      title={The Llama 3 Herd of Models}, 
        author       = {Abhimanyu Dubey and
                  Abhinav Jauhri and
                  Abhinav Pandey and
                  Abhishek Kadian and
                  Ahmad Al{-}Dahle and
                  Aiesha Letman and
                  Akhil Mathur and
                  Alan Schelten and
                  Amy Yang and
                  Angela Fan and
                  Anirudh Goyal and
                  Anthony Hartshorn and
                  Aobo Yang and
                  Archi Mitra and
                  Archie Sravankumar and
                  Artem Korenev and
                  Arthur Hinsvark and
                  Arun Rao and
                  Aston Zhang and
                  Aur{\'{e}}lien Rodriguez and
                  Austen Gregerson and
                  Ava Spataru and
                  Baptiste Rozi{\`{e}}re and
                  Bethany Biron and
                  Binh Tang and
                  Bobbie Chern and
                  Charlotte Caucheteux and
                  Chaya Nayak and
                  Chloe Bi and
                  Chris Marra and
                  Chris McConnell and
                  Christian Keller and
                  Christophe Touret and
                  Chunyang Wu and
                  Corinne Wong and
                  Cristian Canton Ferrer and
                  Cyrus Nikolaidis and
                  Damien Allonsius and
                  Daniel Song and
                  Danielle Pintz and
                  Danny Livshits and
                  David Esiobu and
                  Dhruv Choudhary and
                  Dhruv Mahajan and
                  Diego Garcia{-}Olano and
                  Diego Perino and
                  Dieuwke Hupkes and
                  Egor Lakomkin and
                  Ehab AlBadawy and
                  Elina Lobanova and
                  Emily Dinan and
                  Eric Michael Smith and
                  Filip Radenovic and
                  Frank Zhang and
                  Gabriel Synnaeve and
                  Gabrielle Lee and
                  Georgia Lewis Anderson and
                  Graeme Nail and
                  Gr{\'{e}}goire Mialon and
                  Guan Pang and
                  Guillem Cucurell and
                  Hailey Nguyen and
                  Hannah Korevaar and
                  Hu Xu and
                  Hugo Touvron and
                  Iliyan Zarov and
                  Imanol Arrieta Ibarra and
                  Isabel M. Kloumann and
                  Ishan Misra and
                  Ivan Evtimov and
                  Jade Copet and
                  Jaewon Lee and
                  Jan Geffert and
                  Jana Vranes and
                  Jason Park and
                  Jay Mahadeokar and
                  Jeet Shah and
                  Jelmer van der Linde and
                  Jennifer Billock and
                  Jenny Hong and
                  Jenya Lee and
                  Jeremy Fu and
                  Jianfeng Chi and
                  Jianyu Huang and
                  Jiawen Liu and
                  Jie Wang and
                  Jiecao Yu and
                  Joanna Bitton and
                  Joe Spisak and
                  Jongsoo Park and
                  Joseph Rocca and
                  Joshua Johnstun and
                  Joshua Saxe and
                  Junteng Jia and
                  Kalyan Vasuden Alwala and
                  Kartikeya Upasani and
                  Kate Plawiak and
                  Ke Li and
                  Kenneth Heafield and
                  Kevin Stone and
                  et al.},
      year={2024},
      eprint={2407.21783},
      archivePrefix={arXiv},
      primaryClass={cs.AI},
      url={https://arxiv.org/abs/2407.21783}, 
}

@inproceedings{
flashattention,
title={FlashAttention: Fast and Memory-Efficient Exact Attention with {IO}-Awareness},
author={Tri Dao and Daniel Y Fu and Stefano Ermon and Atri Rudra and Christopher Re},
booktitle={Advances in Neural Information Processing Systems},
editor={Alice H. Oh and Alekh Agarwal and Danielle Belgrave and Kyunghyun Cho},
year={2022},
url={https://openreview.net/forum?id=H4DqfPSibmx}
}

@inproceedings{vllm,
  title={Efficient memory management for large language model serving with pagedattention},
  author={Kwon, Woosuk and Li, Zhuohan and Zhuang, Siyuan and Sheng, Ying and Zheng, Lianmin and Yu, Cody Hao and Gonzalez, Joseph and Zhang, Hao and Stoica, Ion},
  booktitle={Proceedings of the 29th Symposium on Operating Systems Principles},
  pages={611--626},
  year={2023}
}

@inproceedings{xiong-etal-2024-benchmarking,
    title = "Benchmarking Retrieval-Augmented Generation for Medicine",
    author = "Xiong, Guangzhi  and
      Jin, Qiao  and
      Lu, Zhiyong  and
      Zhang, Aidong",
    editor = "Ku, Lun-Wei  and
      Martins, Andre  and
      Srikumar, Vivek",
    booktitle = "Findings of the Association for Computational Linguistics ACL 2024",
    month = aug,
    year = "2024",
    address = "Bangkok, Thailand and virtual meeting",
    publisher = "Association for Computational Linguistics",
    url = "https://aclanthology.org/2024.findings-acl.372",
    pages = "6233--6251",
}

@inproceedings{kalra-etal-2024-hypa,
    title = "{H}y{PA}-{RAG}: A Hybrid Parameter Adaptive Retrieval-Augmented Generation System for {AI} Legal and Policy Applications",
    author = "Kalra, Rishi  and
      Wu, Zekun  and
      Gulley, Ayesha  and
      Hilliard, Airlie  and
      Guan, Xin  and
      Koshiyama, Adriano  and
      Treleaven, Philip Colin",
    editor = "Kumar, Sachin  and
      Balachandran, Vidhisha  and
      Park, Chan Young  and
      Shi, Weijia  and
      Hayati, Shirley Anugrah  and
      Tsvetkov, Yulia  and
      Smith, Noah  and
      Hajishirzi, Hannaneh  and
      Kang, Dongyeop  and
      Jurgens, David",
    booktitle = "Proceedings of the 1st Workshop on Customizable NLP: Progress and Challenges in Customizing NLP for a Domain, Application, Group, or Individual (CustomNLP4U)",
    month = nov,
    year = "2024",
    address = "Miami, Florida, USA",
    publisher = "Association for Computational Linguistics",
    url = "https://aclanthology.org/2024.customnlp4u-1.18/",
    doi = "10.18653/v1/2024.customnlp4u-1.18",
    pages = "237--256"
}

@article{gao2024empowering,
  title={Empowering biomedical discovery with AI agents},
  author={Gao, Shanghua and Fang, Ada and Huang, Yepeng and Giunchiglia, Valentina and Noori, Ayush and Schwarz, Jonathan Richard and Ektefaie, Yasha and Kondic, Jovana and Zitnik, Marinka},
  journal={Cell},
  volume={187},
  number={22},
  pages={6125--6151},
  year={2024},
  publisher={Elsevier}
}

@inproceedings{DBLP:conf/icml/GuuLTPC20,
  author       = {Kelvin Guu and
                  Kenton Lee and
                  Zora Tung and
                  Panupong Pasupat and
                  Ming{-}Wei Chang},
  title        = {Retrieval Augmented Language Model Pre-Training},
  booktitle    = {Proceedings of the 37th International Conference on Machine Learning,
                  {ICML} 2020, 13-18 July 2020, Virtual Event},
  series       = {Proceedings of Machine Learning Research},
  volume       = {119},
  pages        = {3929--3938},
  publisher    = {{PMLR}},
  year         = {2020},
  url          = {http://proceedings.mlr.press/v119/guu20a.html},
  bibsource    = {dblp computer science bibliography, https://dblp.org}
}

@inproceedings{CAG,
      title={Don't Do RAG: When Cache-Augmented Generation is All You Need for Knowledge Tasks}, 
      author={Brian J Chan and Chao-Ting Chen and Jui-Hung Cheng and Hen-Hsen Huang},
  booktitle    = {Companion Proceedings of the {ACM} on Web Conference 2025, {WWW} 2025,
                  Sydney, Australia, Apr 28-May 2, 2025},
  publisher    = {{ACM}},
  year         = {2025},
  bibsource    = {dblp computer science bibliography, https://dblp.org}
}

@inproceedings{DBLP:conf/acm/AsaiMZC23,
  author       = {Akari Asai and
                  Sewon Min and
                  Zexuan Zhong and
                  Danqi Chen},
  editor       = {Yun{-}Nung Vivian Chen and
                  Margot Mieskes and
                  Siva Reddy},
  title        = {Retrieval-based Language Models and Applications},
  booktitle    = {Proceedings of the 61st Annual Meeting of the Association for Computational
                  Linguistics: Tutorial Abstracts, {ACL} 2023, Toronto, Canada, July
                  9-14, 2023},
  pages        = {41--46},
  publisher    = {Association for Computational Linguistics},
  year         = {2023},
  url          = {https://doi.org/10.18653/v1/2023.acl-tutorials.6},
  doi          = {10.18653/V1/2023.ACL-TUTORIALS.6},
  bibsource    = {dblp computer science bibliography, https://dblp.org}
}

@article{DBLP:journals/tacl/RamLDMSLS23,
  author       = {Ori Ram and
                  Yoav Levine and
                  Itay Dalmedigos and
                  Dor Muhlgay and
                  Amnon Shashua and
                  Kevin Leyton{-}Brown and
                  Yoav Shoham},
  title        = {In-Context Retrieval-Augmented Language Models},
  journal      = {Trans. Assoc. Comput. Linguistics},
  volume       = {11},
  pages        = {1316--1331},
  year         = {2023},
  url          = {https://doi.org/10.1162/tacl\_a\_00605},
  doi          = {10.1162/TACL\_A\_00605},
  bibsource    = {dblp computer science bibliography, https://dblp.org}
}

@inproceedings{flashattention2,
  author       = {Tri Dao},
  title        = {FlashAttention-2: Faster Attention with Better Parallelism and Work
                  Partitioning},
  booktitle    = {The Twelfth International Conference on Learning Representations,
                  {ICLR} 2024, Vienna, Austria, May 7-11, 2024},
  publisher    = {OpenReview.net},
  year         = {2024},
  url          = {https://openreview.net/forum?id=mZn2Xyh9Ec},
  bibsource    = {dblp computer science bibliography, https://dblp.org}
}

@misc{h100,
      title={NVIDIA Hopper Architecture In-Depth}, 
      author={NVIDIA},
      year={2022},
      url={https://developer.nvidia.com/blog/nvidia-hopper-architecture-in-depth/}, 
}

@inproceedings{xanh2020_2wikimultihop,
    title = "Constructing A Multi-hop {QA} Dataset for Comprehensive Evaluation of Reasoning Steps",
    author = "Ho, Xanh  and
      Duong Nguyen, Anh-Khoa  and
      Sugawara, Saku  and
      Aizawa, Akiko",
    booktitle = "Proceedings of the 28th International Conference on Computational Linguistics",
    month = dec,
    year = "2020",
    address = "Barcelona, Spain (Online)",
    publisher = "International Committee on Computational Linguistics",
    url = "https://www.aclweb.org/anthology/2020.coling-main.580",
    pages = "6609--6625",
}

@misc{DeepSeekV3,
      title={DeepSeek-V3 Technical Report}, 
      author={DeepSeek-AI and Aixin Liu and Bei Feng and Bing Xue and Bingxuan Wang and Bochao Wu and Chengda Lu and Chenggang Zhao and Chengqi Deng and Chenyu Zhang and Chong Ruan and Damai Dai and Daya Guo and Dejian Yang and Deli Chen and Dongjie Ji and Erhang Li and Fangyun Lin and Fucong Dai and Fuli Luo and Guangbo Hao and Guanting Chen and Guowei Li and H. Zhang and Han Bao and Hanwei Xu and Haocheng Wang and Haowei Zhang and Honghui Ding and Huajian Xin and Huazuo Gao and Hui Li and Hui Qu and J. L. Cai and Jian Liang and Jianzhong Guo and Jiaqi Ni and Jiashi Li and Jiawei Wang and Jin Chen and Jingchang Chen and Jingyang Yuan and Junjie Qiu and Junlong Li and Junxiao Song and Kai Dong and Kai Hu and Kaige Gao and Kang Guan and Kexin Huang and Kuai Yu and Lean Wang and Lecong Zhang and Lei Xu and Leyi Xia and Liang Zhao and Litong Wang and Liyue Zhang and Meng Li and Miaojun Wang and Mingchuan Zhang and Minghua Zhang and Minghui Tang and Mingming Li and Ning Tian and Panpan Huang and Peiyi Wang and Peng Zhang and Qiancheng Wang and Qihao Zhu and Qinyu Chen and Qiushi Du and R. J. Chen and R. L. Jin and Ruiqi Ge and Ruisong Zhang and Ruizhe Pan and Runji Wang and Runxin Xu and Ruoyu Zhang and Ruyi Chen and S. S. Li and Shanghao Lu and Shangyan Zhou and Shanhuang Chen and Shaoqing Wu and Shengfeng Ye and Shengfeng Ye and Shirong Ma and Shiyu Wang and Shuang Zhou and Shuiping Yu and Shunfeng Zhou and Shuting Pan and T. Wang and Tao Yun and Tian Pei and Tianyu Sun and W. L. Xiao and Wangding Zeng and Wanjia Zhao and Wei An and Wen Liu and Wenfeng Liang and Wenjun Gao and Wenqin Yu and Wentao Zhang and X. Q. Li and Xiangyue Jin and Xianzu Wang and Xiao Bi and Xiaodong Liu and Xiaohan Wang and Xiaojin Shen and Xiaokang Chen and Xiaokang Zhang and Xiaosha Chen and Xiaotao Nie and Xiaowen Sun and Xiaoxiang Wang and Xin Cheng and Xin Liu and Xin Xie and Xingchao Liu and Xingkai Yu and Xinnan Song and Xinxia Shan and Xinyi Zhou and Xinyu Yang and Xinyuan Li and Xuecheng Su and Xuheng Lin and Y. K. Li and Y. Q. Wang and Y. X. Wei and Y. X. Zhu and Yang Zhang and Yanhong Xu and Yanhong Xu and Yanping Huang and Yao Li and Yao Zhao and Yaofeng Sun and Yaohui Li and Yaohui Wang and Yi Yu and Yi Zheng and Yichao Zhang and Yifan Shi and Yiliang Xiong and Ying He and Ying Tang and Yishi Piao and Yisong Wang and Yixuan Tan and Yiyang Ma and Yiyuan Liu and Yongqiang Guo and Yu Wu and Yuan Ou and Yuchen Zhu and Yuduan Wang and Yue Gong and Yuheng Zou and Yujia He and Yukun Zha and Yunfan Xiong and Yunxian Ma and Yuting Yan and Yuxiang Luo and Yuxiang You and Yuxuan Liu and Yuyang Zhou and Z. F. Wu and Z. Z. Ren and Zehui Ren and Zhangli Sha and Zhe Fu and Zhean Xu and Zhen Huang and Zhen Zhang and Zhenda Xie and Zhengyan Zhang and Zhewen Hao and Zhibin Gou and Zhicheng Ma and Zhigang Yan and Zhihong Shao and Zhipeng Xu and Zhiyu Wu and Zhongyu Zhang and Zhuoshu Li and Zihui Gu and Zijia Zhu and Zijun Liu and Zilin Li and Ziwei Xie and Ziyang Song and Ziyi Gao and Zizheng Pan},
      year={2025},
      eprint={2412.19437},
      archivePrefix={arXiv},
      primaryClass={cs.CL},
      url={https://arxiv.org/abs/2412.19437}, 
}

@inproceedings{
meng2025transmla,
title={Trans{MLA}: Migrating {GQA} Models to {MLA} with Full DeepSeek Compatibility and Speedup},
author={Fanxu Meng and Pingzhi Tang and Zengwei Yao and Xing Sun and Muhan Zhang},
booktitle={The Thirty-ninth Annual Conference on Neural Information Processing Systems},
year={2025},
url={https://openreview.net/forum?id=TcVCu2PKb9}
}

@article{DeepSeekV2,
  author       = {DeepSeek{-}AI and
                  Aixin Liu and
                  Bei Feng and
                  Bin Wang and
                  Bingxuan Wang and
                  Bo Liu and
                  Chenggang Zhao and
                  Chengqi Deng and
                  Chong Ruan and
                  Damai Dai and
                  Daya Guo and
                  Dejian Yang and
                  Deli Chen and
                  Dongjie Ji and
                  Erhang Li and
                  Fangyun Lin and
                  Fuli Luo and
                  Guangbo Hao and
                  Guanting Chen and
                  Guowei Li and
                  Hao Zhang and
                  Hanwei Xu and
                  Hao Yang and
                  Haowei Zhang and
                  Honghui Ding and
                  Huajian Xin and
                  Huazuo Gao and
                  Hui Li and
                  Hui Qu and
                  J. L. Cai and
                  Jian Liang and
                  Jianzhong Guo and
                  Jiaqi Ni and
                  Jiashi Li and
                  Jin Chen and
                  Jingyang Yuan and
                  Junjie Qiu and
                  Junxiao Song and
                  Kai Dong and
                  Kaige Gao and
                  Kang Guan and
                  Lean Wang and
                  Lecong Zhang and
                  Lei Xu and
                  Leyi Xia and
                  Liang Zhao and
                  Liyue Zhang and
                  Meng Li and
                  Miaojun Wang and
                  Mingchuan Zhang and
                  Minghua Zhang and
                  Minghui Tang and
                  Mingming Li and
                  Ning Tian and
                  Panpan Huang and
                  Peiyi Wang and
                  Peng Zhang and
                  Qihao Zhu and
                  Qinyu Chen and
                  Qiushi Du and
                  R. J. Chen and
                  R. L. Jin and
                  Ruiqi Ge and
                  Ruizhe Pan and
                  Runxin Xu and
                  Ruyi Chen and
                  S. S. Li and
                  Shanghao Lu and
                  Shangyan Zhou and
                  Shanhuang Chen and
                  Shaoqing Wu and
                  Shengfeng Ye and
                  Shirong Ma and
                  Shiyu Wang and
                  Shuang Zhou and
                  Shuiping Yu and
                  Shunfeng Zhou and
                  Size Zheng and
                  Tao Wang and
                  Tian Pei and
                  Tian Yuan and
                  Tianyu Sun and
                  W. L. Xiao and
                  Wangding Zeng and
                  Wei An and
                  Wen Liu and
                  Wenfeng Liang and
                  Wenjun Gao and
                  Wentao Zhang and
                  X. Q. Li and
                  Xiangyue Jin and
                  Xianzu Wang and
                  Xiao Bi and
                  Xiaodong Liu and
                  Xiaohan Wang and
                  Xiaojin Shen and
                  Xiaokang Chen and
                  Xiaosha Chen and
                  Xiaotao Nie and
                  Xiaowen Sun},
  title        = {DeepSeek-V2: {A} Strong, Economical, and Efficient Mixture-of-Experts
                  Language Model},
  journal      = {CoRR},
  volume       = {abs/2405.04434},
  year         = {2024},
  url          = {https://doi.org/10.48550/arXiv.2405.04434},
  doi          = {10.48550/ARXIV.2405.04434},
  eprinttype    = {arXiv},
  eprint       = {2405.04434},
  bibsource    = {dblp computer science bibliography, https://dblp.org}
}

@inproceedings{DBLP:conf/acl/JoshiCWZ17,
  author       = {Mandar Joshi and
                  Eunsol Choi and
                  Daniel S. Weld and
                  Luke Zettlemoyer},
  editor       = {Regina Barzilay and
                  Min{-}Yen Kan},
  title        = {TriviaQA: {A} Large Scale Distantly Supervised Challenge Dataset for
                  Reading Comprehension},
  booktitle    = {Proceedings of the 55th Annual Meeting of the Association for Computational
                  Linguistics, {ACL} 2017, Vancouver, Canada, July 30 - August 4, Volume
                  1: Long Papers},
  pages        = {1601--1611},
  publisher    = {Association for Computational Linguistics},
  year         = {2017},
  url          = {https://doi.org/10.18653/v1/P17-1147},
  doi          = {10.18653/V1/P17-1147},
  bibsource    = {dblp computer science bibliography, https://dblp.org}
}

@article{kocisky-etal-2018-narrativeqa,
    title = "The {N}arrative{QA} Reading Comprehension Challenge",
    author = "Ko{\v{c}}isk{\'y}, Tom{\'a}{\v{s}}  and
      Schwarz, Jonathan  and
      Blunsom, Phil  and
      Dyer, Chris  and
      Hermann, Karl Moritz  and
      Melis, G{\'a}bor  and
      Grefenstette, Edward",
    editor = "Lee, Lillian  and
      Johnson, Mark  and
      Toutanova, Kristina  and
      Roark, Brian",
    journal = "Transactions of the Association for Computational Linguistics",
    volume = "6",
    year = "2018",
    address = "Cambridge, MA",
    publisher = "MIT Press",
    url = "https://aclanthology.org/Q18-1023/",
    doi = "10.1162/tacl_a_00023",
    pages = "317--328"
}

@inproceedings{yang-etal-2018-hotpotqa,
    title = "{H}otpot{QA}: A Dataset for Diverse, Explainable Multi-hop Question Answering",
    author = "Yang, Zhilin  and
      Qi, Peng  and
      Zhang, Saizheng  and
      Bengio, Yoshua  and
      Cohen, William  and
      Salakhutdinov, Ruslan  and
      Manning, Christopher D.",
    editor = "Riloff, Ellen  and
      Chiang, David  and
      Hockenmaier, Julia  and
      Tsujii, Jun{'}ichi",
    booktitle = "Proceedings of the 2018 Conference on Empirical Methods in Natural Language Processing",
    month = oct # "-" # nov,
    year = "2018",
    address = "Brussels, Belgium",
    publisher = "Association for Computational Linguistics",
    url = "https://aclanthology.org/D18-1259/",
    doi = "10.18653/v1/D18-1259",
    pages = "2369--2380"
}

@inproceedings{DBLP:conf/eacl/IzacardG21,
  author       = {Gautier Izacard and
                  Edouard Grave},
  editor       = {Paola Merlo and
                  J{\"{o}}rg Tiedemann and
                  Reut Tsarfaty},
  title        = {Leveraging Passage Retrieval with Generative Models for Open Domain
                  Question Answering},
  booktitle    = {Proceedings of the 16th Conference of the European Chapter of the
                  Association for Computational Linguistics: Main Volume, {EACL} 2021,
                  Online, April 19 - 23, 2021},
  pages        = {874--880},
  publisher    = {Association for Computational Linguistics},
  year         = {2021},
  url          = {https://doi.org/10.18653/v1/2021.eacl-main.74},
  doi          = {10.18653/V1/2021.EACL-MAIN.74},
  bibsource    = {dblp computer science bibliography, https://dblp.org}
}

@inproceedings{DBLP:conf/icml/BorgeaudMHCRM0L22,
  author       = {Sebastian Borgeaud and
                  Arthur Mensch and
                  Jordan Hoffmann and
                  Trevor Cai and
                  Eliza Rutherford and
                  Katie Millican and
                  George van den Driessche and
                  Jean{-}Baptiste Lespiau and
                  Bogdan Damoc and
                  Aidan Clark and
                  Diego de Las Casas and
                  Aurelia Guy and
                  Jacob Menick and
                  Roman Ring and
                  Tom Hennigan and
                  Saffron Huang and
                  Loren Maggiore and
                  Chris Jones and
                  Albin Cassirer and
                  Andy Brock and
                  Michela Paganini and
                  Geoffrey Irving and
                  Oriol Vinyals and
                  Simon Osindero and
                  Karen Simonyan and
                  Jack W. Rae and
                  Erich Elsen and
                  Laurent Sifre},
  editor       = {Kamalika Chaudhuri and
                  Stefanie Jegelka and
                  Le Song and
                  Csaba Szepesv{\'{a}}ri and
                  Gang Niu and
                  Sivan Sabato},
  title        = {Improving Language Models by Retrieving from Trillions of Tokens},
  booktitle    = {International Conference on Machine Learning, {ICML} 2022, 17-23 July
                  2022, Baltimore, Maryland, {USA}},
  series       = {Proceedings of Machine Learning Research},
  volume       = {162},
  pages        = {2206--2240},
  publisher    = {{PMLR}},
  year         = {2022},
  url          = {https://proceedings.mlr.press/v162/borgeaud22a.html},
  bibsource    = {dblp computer science bibliography, https://dblp.org}
}

@misc{vLLMv1,
      title={{vLLM V1}: A Major Upgrade to {vLLM}'s Core Architecture}, 
      author={{vLLM Team}},
      year={2025},
      url={https://blog.vllm.ai/2025/01/27/v1-alpha-release.html}, 
}

@inproceedings{DBLP:conf/acl/LiRZWLVYK23,
  author       = {Daliang Li and
                  Ankit Singh Rawat and
                  Manzil Zaheer and
                  Xin Wang and
                  Michal Lukasik and
                  Andreas Veit and
                  Felix X. Yu and
                  Sanjiv Kumar},
  editor       = {Anna Rogers and
                  Jordan L. Boyd{-}Graber and
                  Naoaki Okazaki},
  title        = {Large Language Models with Controllable Working Memory},
  booktitle    = {Findings of the Association for Computational Linguistics: {ACL} 2023,
                  Toronto, Canada, July 9-14, 2023},
  pages        = {1774--1793},
  publisher    = {Association for Computational Linguistics},
  year         = {2023},
  url          = {https://doi.org/10.18653/v1/2023.findings-acl.112},
  doi          = {10.18653/V1/2023.FINDINGS-ACL.112},
  bibsource    = {dblp computer science bibliography, https://dblp.org}
}

@inproceedings{DBLP:conf/iclr/YoranWRB24,
  author       = {Ori Yoran and
                  Tomer Wolfson and
                  Ori Ram and
                  Jonathan Berant},
  title        = {Making Retrieval-Augmented Language Models Robust to Irrelevant Context},
  booktitle    = {The Twelfth International Conference on Learning Representations,
                  {ICLR} 2024, Vienna, Austria, May 7-11, 2024},
  publisher    = {OpenReview.net},
  year         = {2024},
  url          = {https://openreview.net/forum?id=ZS4m74kZpH},
  bibsource    = {dblp computer science bibliography, https://dblp.org}
}

@inproceedings{DBLP:conf/sigir/CuconasuTSFCMTS24,
  author       = {Florin Cuconasu and
                  Giovanni Trappolini and
                  Federico Siciliano and
                  Simone Filice and
                  Cesare Campagnano and
                  Yoelle Maarek and
                  Nicola Tonellotto and
                  Fabrizio Silvestri},
  editor       = {Grace Hui Yang and
                  Hongning Wang and
                  Sam Han and
                  Claudia Hauff and
                  Guido Zuccon and
                  Yi Zhang},
  title        = {The Power of Noise: Redefining Retrieval for {RAG} Systems},
  booktitle    = {Proceedings of the 47th International {ACM} {SIGIR} Conference on
                  Research and Development in Information Retrieval, {SIGIR} 2024, Washington
                  DC, USA, July 14-18, 2024},
  pages        = {719--729},
  publisher    = {{ACM}},
  year         = {2024},
  url          = {https://doi.org/10.1145/3626772.3657834},
  doi          = {10.1145/3626772.3657834},
  bibsource    = {dblp computer science bibliography, https://dblp.org}
}

@article{RoPE,
  author       = {Jianlin Su and
                  Murtadha H. M. Ahmed and
                  Yu Lu and
                  Shengfeng Pan and
                  Wen Bo and
                  Yunfeng Liu},
  title        = {RoFormer: Enhanced transformer with Rotary Position Embedding},
  journal      = {Neurocomputing},
  volume       = {568},
  pages        = {127063},
  year         = {2024},
  url          = {https://doi.org/10.1016/j.neucom.2023.127063},
  doi          = {10.1016/J.NEUCOM.2023.127063},
  bibsource    = {dblp computer science bibliography, https://dblp.org}
}

@inproceedings{
jiang2024minference,
title={{MI}nference 1.0: Accelerating Pre-filling for Long-Context {LLM}s via Dynamic Sparse Attention},
author={Huiqiang Jiang and YUCHENG LI and Chengruidong Zhang and Qianhui Wu and Xufang Luo and Surin Ahn and Zhenhua Han and Amir H. Abdi and Dongsheng Li and Chin-Yew Lin and Yuqing Yang and Lili Qiu},
booktitle={The Thirty-eighth Annual Conference on Neural Information Processing Systems},
year={2024},
url={https://openreview.net/forum?id=fPBACAbqSN}
}

@inproceedings{
zheng2024sglang,
title={{SGL}ang: Efficient Execution of Structured Language Model Programs},
author={Lianmin Zheng and Liangsheng Yin and Zhiqiang Xie and Chuyue Sun and Jeff Huang and Cody Hao Yu and Shiyi Cao and Christos Kozyrakis and Ion Stoica and Joseph E. Gonzalez and Clark Barrett and Ying Sheng},
booktitle={The Thirty-eighth Annual Conference on Neural Information Processing Systems},
year={2024},
url={https://openreview.net/forum?id=VqkAKQibpq}
}

@inproceedings{triton,
author = {Tillet, Philippe and Kung, H. T. and Cox, David},
title = {Triton: an intermediate language and compiler for tiled neural network computations},
year = {2019},
isbn = {9781450367196},
publisher = {Association for Computing Machinery},
address = {New York, NY, USA},
url = {https://doi.org/10.1145/3315508.3329973},
doi = {10.1145/3315508.3329973},
booktitle = {Proceedings of the 3rd ACM SIGPLAN International Workshop on Machine Learning and Programming Languages},
pages = {10–19},
numpages = {10},
location = {Phoenix, AZ, USA},
series = {MAPL 2019}
}

@inproceedings{
flexattention,
title={FlexAttention: A Programming Model for Generating Fused Attention Variants.},
author={Juechu Dong and BOYUAN FENG and Driss Guessous and Yanbo Liang and Horace He},
booktitle={Eighth Conference on Machine Learning and Systems},
year={2025},
url={https://openreview.net/forum?id=2QMYV4bA0R}
}

@inproceedings{transformer,
  author       = {Ashish Vaswani and
                  Noam Shazeer and
                  Niki Parmar and
                  Jakob Uszkoreit and
                  Llion Jones and
                  Aidan N. Gomez and
                  Lukasz Kaiser and
                  Illia Polosukhin},
  editor       = {Isabelle Guyon and
                  Ulrike von Luxburg and
                  Samy Bengio and
                  Hanna M. Wallach and
                  Rob Fergus and
                  S. V. N. Vishwanathan and
                  Roman Garnett},
  title        = {Attention is All you Need},
  booktitle    = {Advances in Neural Information Processing Systems 30: Annual Conference
                  on Neural Information Processing Systems 2017, December 4-9, 2017,
                  Long Beach, CA, {USA}},
  pages        = {5998--6008},
  year         = {2017},
  url          = {https://proceedings.neurips.cc/paper/2017/hash/3f5ee243547dee91fbd053c1c4a845aa-Abstract.html},
  bibsource    = {dblp computer science bibliography, https://dblp.org}
}

@inproceedings{xlnet,
  author       = {Zhilin Yang and
                  Zihang Dai and
                  Yiming Yang and
                  Jaime G. Carbonell and
                  Ruslan Salakhutdinov and
                  Quoc V. Le},
  editor       = {Hanna M. Wallach and
                  Hugo Larochelle and
                  Alina Beygelzimer and
                  Florence d'Alch{\'{e}}{-}Buc and
                  Emily B. Fox and
                  Roman Garnett},
  title        = {XLNet: Generalized Autoregressive Pretraining for Language Understanding},
  booktitle    = {Advances in Neural Information Processing Systems 32: Annual Conference
                  on Neural Information Processing Systems 2019, NeurIPS 2019, December
                  8-14, 2019, Vancouver, BC, Canada},
  pages        = {5754--5764},
  year         = {2019},
  url          = {https://proceedings.neurips.cc/paper/2019/hash/dc6a7e655d7e5840e66733e9ee67cc69-Abstract.html},
  bibsource    = {dblp computer science bibliography, https://dblp.org}
}

@misc{llama4,
      title={The Llama 4 herd: The beginning of a new era of natively multimodal AI innovation}, 
      author={{Llama Team}},
      year={2025},
      url={https://ai.meta.com/blog/llama-4-multimodal-intelligence/}, 
}

@misc{llama3.1,
      title={Introducing Llama 3.1: Our most capable models to date}, 
      author={{Llama Team}},
      year={2024},
      url={https://ai.meta.com/blog/meta-llama-3-1/}, 
}

@inproceedings{rope-scaling,
  author       = {Xiaoran Liu and
                  Hang Yan and
                  Chenxin An and
                  Xipeng Qiu and
                  Dahua Lin},
  title        = {Scaling Laws of RoPE-based Extrapolation},
  booktitle    = {The Twelfth International Conference on Learning Representations,
                  {ICLR} 2024, Vienna, Austria, May 7-11, 2024},
  publisher    = {OpenReview.net},
  year         = {2024},
  url          = {https://openreview.net/forum?id=JO7k0SJ5V6},
  bibsource    = {dblp computer science bibliography, https://dblp.org}
}

@inproceedings{
flashattention3,
title={FlashAttention-3: Fast and Accurate Attention with Asynchrony and Low-precision},
author={Jay Shah and Ganesh Bikshandi and Ying Zhang and Vijay Thakkar and Pradeep Ramani and Tri Dao},
booktitle={The Thirty-eighth Annual Conference on Neural Information Processing Systems},
year={2024},
url={https://openreview.net/forum?id=tVConYid20}
}

@inproceedings{DBLP:conf/icml/TangZZXKH24,
  author       = {Jiaming Tang and
                  Yilong Zhao and
                  Kan Zhu and
                  Guangxuan Xiao and
                  Baris Kasikci and
                  Song Han},
  title        = {{QUEST:} Query-Aware Sparsity for Efficient Long-Context {LLM} Inference},
  booktitle    = {Forty-first International Conference on Machine Learning, {ICML} 2024,
                  Vienna, Austria, July 21-27, 2024},
  publisher    = {OpenReview.net},
  year         = {2024},
  url          = {https://openreview.net/forum?id=KzACYw0MTV},
  bibsource    = {dblp computer science bibliography, https://dblp.org}
}

@inproceedings{DBLP:conf/icml/AcharyaJG25,
  author       = {Shantanu Acharya and
                  Fei Jia and
                  Boris Ginsburg},
  title        = {Star Attention: Efficient {LLM} Inference over Long Sequences},
  booktitle    = {Forty-second International Conference on Machine Learning, {ICML}
                  2025, Vancouver, BC, Canada, July 13-19, 2025},
  publisher    = {OpenReview.net},
  year         = {2025},
  url          = {https://openreview.net/forum?id=QY7Au9nZwp},
  bibsource    = {dblp computer science bibliography, https://dblp.org}
}

@inproceedings{DBLP:conf/icml/ZhangXHWX0C25,
  author       = {Jintao Zhang and
                  Chendong Xiang and
                  Haofeng Huang and
                  Jia Wei and
                  Haocheng Xi and
                  Jun Zhu and
                  Jianfei Chen},
  title        = {SpargeAttention: Accurate and Training-free Sparse Attention Accelerating
                  Any Model Inference},
  booktitle    = {Forty-second International Conference on Machine Learning, {ICML}
                  2025, Vancouver, BC, Canada, July 13-19, 2025},
  publisher    = {OpenReview.net},
  year         = {2025},
  url          = {https://openreview.net/forum?id=74c3Wwk8Tc},
  bibsource    = {dblp computer science bibliography, https://dblp.org}
}

@inproceedings{DBLP:conf/icml/QinS0SSZ24,
  author       = {Zhen Qin and
                  Weigao Sun and
                  Dong Li and
                  Xuyang Shen and
                  Weixuan Sun and
                  Yiran Zhong},
  title        = {Various Lengths, Constant Speed: Efficient Language Modeling with
                  Lightning Attention},
  booktitle    = {Forty-first International Conference on Machine Learning, {ICML} 2024,
                  Vienna, Austria, July 21-27, 2024},
  publisher    = {OpenReview.net},
  year         = {2024},
  url          = {https://openreview.net/forum?id=Lwm6TiUP4X},
  bibsource    = {dblp computer science bibliography, https://dblp.org}
}

@inproceedings{alibi2022,
  title={Train Short, Test Long: Attention with Linear Biases Enables Input Length Extrapolation},
  author={Press, Ofir and Smith, Noah A. and Lewis, Mike},
  booktitle={International Conference on Learning Representations},
  year={2022},
  url={https://openreview.net/forum?id=R8sQPpGCv0}
}

@article{falcon2023,
  title={The Falcon Series of Open Language Models},
  author={Almazrouei, Ebtesam and Alobeidli, Hamza and Alshamsi, Abdulaziz and Cappelli, Alessandro and Cojocaru, Ruxandra and Debbah, Merouane and Goffinet, Etienne and Hesslow, Daniel and Launay, Julien and Malartic, Quentin and Mazzotta, Daniele and Noune, Badreddine and Pannier, Baptiste and Penedo, Guilherme},
  journal={arXiv preprint arXiv:2311.16867},
  year={2023},
  url={https://arxiv.org/abs/2311.16867}
}

@inproceedings{agnews2015,
  title={Character-level Convolutional Networks for Text Classification},
  author={Zhang, Xiang and Zhao, Junbo and LeCun, Yann},
  booktitle={Advances in Neural Information Processing Systems},
  year={2015},
  url={https://proceedings.neurips.cc/paper/2015/hash/250cf8b51c773f3f8dc8b4be867a9a02-Abstract.html}
}

@misc{mepic,
      title={MEPIC: Memory Efficient Position Independent Caching for LLM Serving}, 
      author={Qian Wang and Zahra Yousefijamarani and Morgan Lindsay Heisler and Rongzhi Gu and Bai Xiaolong and Shan Yizhou and Wei Zhang and Wang Lan and Ying Xiong and Yong Zhang and Zhenan Fan},
      year={2025},
      eprint={2512.16822},
      archivePrefix={arXiv},
      primaryClass={cs.LG},
      url={https://arxiv.org/abs/2512.16822}, 
}
\bibliographystyle{icml2026}

\newpage
\appendix
\onecolumn


\section{Related Work}
We organize related work around three system questions: what cache units are kept, how positional information is handled when those units move, and how attention kernels reduce memory traffic.

\paragraph{KV reuse and cache management}
Serving systems such as RAGCache~\citep{DBLP:journals/corr/abs-2404-12457} and Mooncake~\citep{DBLP:journals/corr/abs-2407-00079} improve cache hierarchy, eviction, and transfer policies, but the reusable unit remains a standard position-aware KV prefix. As a result, a document can be reused only when it appears in the same contiguous prefix layout; otherwise the system must recompute the states or store another copy. Prompt Cache~\citep{DBLP:conf/mlsys/GimCLSK024}, CacheBlend~\citep{DBLP:conf/eurosys/cacheblend}, TurboRAG~\citep{lu-etal-2025-turborag}, and Block-Attention~\citep{ma2025blockattention} broaden reuse by matching chunks or re-encoding positional information, but still pay either recomputation, reconstruction, or materialization costs. \ourmethod{} is complementary to cache-management policies: they decide \emph{what} to keep, while our mechanism changes \emph{how} cached states are consumed so that one physical KV block can serve multiple logical positions.

\paragraph{Decoupling position from stored KV states}
Several recent directions decouple positional information from stored representations, but with different goals and costs. MLA-style architectures such as DeepSeek-V2/V3~\citep{DeepSeekV2,DeepSeekV3} and TransMLA~\citep{meng2025transmla} use decoupled RoPE primarily for KV compression through low-rank representations, not for arbitrary-position document reuse. Position-independent caching methods such as EPIC~\citep{hu2025epic}, MEPIC-like~\cite{mepic} fused-position designs, KVShare, CacheClip, CacheSlide, and KVLink target more flexible reuse, while pre-RoPE-key techniques such as KVQuant, ShadowKV, and XQuant use related representation choices mainly for compression or quantization.
The key distinction is where the positional adjustment is paid. Methods that materialize shifted KV states consume extra HBM capacity and bandwidth; methods that require partial recomputation lose the benefit of reuse. \ourmethod{} instead keeps document-level offsets as metadata and injects the relative positional effect inside the attention kernel, avoiding materialized shifted copies while preserving exact attention for identical cached chunks. As concurrent work, MEPIC~\cite{mepic} shares a similar design principle with \ourmethod. However, MEPIC introduces additional I/O operations because it applies different positional rotations per token, whereas \ourmethod shifts all tokens in a document by the same offset.

\paragraph{IO-aware attention kernels}
FlashAttention-2/3~\citep{flashattention2,flashattention3}, FlexAttention~\citep{flexattention}, and Lightning Attention~\citep{DBLP:conf/icml/QinS0SSZ24} show that attention performance is often governed by memory traffic and kernel fusion rather than FLOPs alone. \ourmethod{} follows the same IO-aware principle, but applies it to RAG reuse: in prefilling, deferred rotation adds only a small per-tile computation and minimal metadata reads; in decoding, packed metadata and in-register rotation avoid extra inner-loop memory traffic. Thus, our contribution is not a new cache policy or a new model architecture, but a kernel-level realization of position-agnostic reuse that preserves the throughput benefits of modern attention kernels.

\section{Experiments Details}~\label{appx:exp}

Here we list the details of the methods compared in the paper.

\begin{itemize}
    \item \textbf{Full Recomputation}: vLLM with \texttt{enable\_prefix\_caching=False}.
    \item \textbf{Prefix Caching}: vLLM with \texttt{enable\_prefix\_caching=True}.
    \item \textbf{Full Reuse}: The KV caches from individual documents are concatenated and used directly as the KV cache for the concatenated documents.
    \item \textbf{Block-Attention (Official)}~\citep{ma2025blockattention}: Evaluated using the official implementation\footnote{\url{https://github.com/TemporaryLoRA/Block-Attention}}.
    \item \textbf{Block-Attention (vLLM)}: We reimplemented Block-Attention~\citep{ma2025blockattention} within the vLLM v1 engine. If a document is found in the cache but its position does not align and its reference count is zero (i.e., no other request is using it), rotation can be applied directly to adjust the positional encoding. Otherwise, new memory is allocated, the blocks are copied, and then rotated.
    \item \textbf{CacheBlend}~\citep{DBLP:conf/eurosys/cacheblend}: Evaluated using the official LMCache examples\footnote{\url{https://github.com/LMCache/LMCache}}. Multiple documents are concatenated with \texttt{blend\_special\_str} to form the prompt, which is then processed by a vLLM instance integrated with LMCache.
    \item \textbf{Prompt Cache}~\citep{DBLP:conf/mlsys/GimCLSK024}: Evaluated using the official implementation\footnote{\url{https://github.com/yale-sys/prompt-cache}}.
\end{itemize}




{
\section{Additional Experimental Results}\label{appx:additional_exp}

\subsection{Generalization to Larger Models}~\label{appx:largermodel}
We evaluate \ourmethod on Llama-3.1-70B-Instruct deployed on a node with 4$\times$H100 GPUs (Tensor Parallelism = 4), under the same RAG QA workload as Section 4.1 (5 retrieved documents, 1 request/s).
Table~\ref{tab:70b_results} shows that \ourmethod achieves a 5.2$\times$ TTFT speedup over standard RAG serving.
Moreover, the gap between \ourmethod and CacheBlend widens from 1.43$\times$ on 8B to 1.53$\times$ on 70B, because larger models have substantially larger KV states (more layers and wider dimensions) and are therefore more memory-bandwidth bound. By avoiding materialization of position-shifted KV blocks, our zero-copy design saves HBM bandwidth, and this advantage amplifies as model size increases.

\begin{table}[h]
\centering
\caption{TTFT on RAG QA for 8B vs 70B models.}
\label{tab:70b_results}
\begin{tabular}{llcc}
\toprule
Model & Method & TTFT (ms) & Speedup \\
\midrule
Tulu3-Block-FT 8B & Standard RAG & 1196.8 & 1.0$\times$ \\
& CacheBlend & 274.8 & 4.4$\times$ \\
& Lazy-Attn (ours) & {191.7} & {6.2$\times$} \\
\midrule
Llama-3.1-70B (TP=4) & Standard RAG & 1253.0 & 1.0$\times$ \\
& CacheBlend & 365.5 & 3.4$\times$ \\
& Lazy-Attn (ours) & {238.4} & {5.2$\times$} \\
\bottomrule
\end{tabular}
\end{table}

\subsection{Performance on Different Hardware}
To test robustness across hardware generations, we further evaluate \ourmethod on NVIDIA A100 (40GB) and A40 (48GB) GPUs, which provide substantially lower memory bandwidth than H100.
As summarized in Table~\ref{tab:hardware_results}, the relative speedup of \ourmethod over CacheBlend grows as memory bandwidth decreases (from 1.43$\times$ on H100 to 1.70$\times$ on A40).
This trend follows from the fact that Block-Attention and CacheBlend use a ``Read-Modify-Write" pattern to materialize position-adjusted KV blocks, effectively doubling HBM traffic. On bandwidth-constrained GPUs like A40, this extra traffic becomes the bottleneck. In contrast, \ourmethod keeps KV accesses strictly read-only and avoids this bandwidth penalty.

\begin{table}[t]
\centering
\caption{TTFT Comparison under Uniform Sampled Documents across different GPUs.}
\label{tab:hardware_results}
\begin{tabular}{llcc}
\toprule
Hardware & Method & TTFT (ms) & Speedup (vs Prefix) \\
\midrule
H100 & Prefix Caching & 1196.8 & 1.0$\times$ \\
(High BW) & CacheBlend & 274.8 & 4.3$\times$ \\
& Lazy-Attn (ours) & {191.7} & {6.2$\times$} \\
\midrule
A100 & Prefix Caching & 2580.4 & 1.0$\times$ \\
(Mid BW) & CacheBlend & 565.3 & 4.5$\times$ \\
& Lazy-Attn (ours) & {372.5} & {6.9$\times$} \\
\midrule
A40 & Prefix Caching & 4150.5 & 1.0$\times$ \\
(Low BW) & CacheBlend & 1150.6 & 3.6$\times$ \\
& Lazy-Attn (ours) & {675.2} & {6.1$\times$} \\
\bottomrule
\end{tabular}
\end{table}

\subsection{Versatility in Tasks: Long-form Literature Review}~\label{appx:lit_review}
To assess versatility beyond factoid-style QA, we construct a Long-form Literature Review task: the model receives 5 ArXiv papers (converted to text, $\sim$8K tokens each) and is asked to produce a 1024-token literature review.
This setting stresses the system on both long-context prefilling and subsequent decoding.
As shown in Table~\ref{tab:lit_review}, \ourmethod improves end-to-end latency by 1.7$\times$, indicating that the benefits of our design extend from TTFT to overall response time in more realistic generative workloads.

\begin{table}[t]
\centering
\caption{Long-form literature review latency (Tulu3-Block-FT 8B, H100).}
\label{tab:lit_review}
\begin{tabular}{lcc}
\toprule
Method & End-to-end Latency (s) & Speedup \\
\midrule
Standard RAG & 38.0 & 1.0$\times$ \\
Lazy-Attn (ours) & {22.4} & {1.7$\times$} \\
\bottomrule
\end{tabular}
\end{table}

\subsection{Long-Context Scalability and Numerical Stability}~\label{appendix:scaling-length}
We next study scalability to longer documents (up to 16K tokens per document) and numerical stability at very long sequence lengths (up to 128K tokens).
Table~\ref{tab:long_doc_scaling} shows that the TTFT speedup of \ourmethod remains roughly constant as document length increases, indicating that our fixed tiling strategy stays efficient even for long contexts.
For stability, \ourmethod applies stateless on-the-fly rotations using absolute token indices ($m\theta_i$), instead of iteratively updating the rotation state. This design prevents error accumulation across layers or timesteps.
Table~\ref{tab:stability} shows that the maximum difference in attention logits versus standard attention remains below 10$^{-5}$ even at 128K tokens, confirming that our kernel is numerically stable.

\begin{table}[t]
\centering
\caption{Performance scaling on varying document lengths (5 docs per request).}
\label{tab:long_doc_scaling}
\begin{tabular}{cccc}
\toprule
Doc length & Standard TTFT & Lazy-Attn TTFT & Speedup \\
\midrule
4K & 7.152 s & 1.487 s & 4.81$\times$ \\
8K & 14.133 s & 2.919 s & 4.84$\times$ \\
16K & 28.446 s & 5.721 s & 4.97$\times$ \\
\bottomrule
\end{tabular}
\end{table}

\begin{table}[t]
\centering
\caption{Consistency with standard attention (H100, 128K tokens).}
\label{tab:stability}
\begin{tabular}{ccc}
\toprule
Seq Length & Max abs diff (logits) & Max relative diff (logits) \\
\midrule
128K & 3.7457e-5 & 5.4829e-7 \\
\bottomrule
\end{tabular}
\end{table}

\subsection{Generalization to Other Architectures and Methods}
We also apply \ourmethod to {Qwen3-8B}~\citep{yang2025qwen3technicalreport}, which uses a different RoPE implementation, and combine it with {Lego-Link0}~\citep{hu2025epic}, a training-free cache reuse strategy.
Table~\ref{tab:generalization_arch} reports consistent speedups of about 6.3$\times$ across these settings, indicating that our kernel is robust to architectural changes and is complementary to both fine-tuned and training-free reuse methods.

\begin{table}[t]
\centering
\caption{Generalization on Qwen3-8B and Lego-Link0.}
\label{tab:generalization_arch}
\begin{tabular}{lccc}
\toprule
Model / Method & Standard TTFT & Lazy-Attn TTFT & Speedup \\
\midrule
Tulu3-Block-FT 8B & 1196.8 ms & 191.7 ms & 6.2$\times$ \\
Qwen3-8B & 1277.4 ms & 201.2 ms & 6.35$\times$ \\
Lego-Link0 + Llama-3.1-8B & 1195.6 ms & 191.3 ms & 6.2$\times$ \\
\bottomrule
\end{tabular}
\end{table}

\subsection{Sensitivity Analysis of Tiling Parameters}
We perform a sensitivity study over different prefilling tile sizes $M$ while fixing the decode tile size at $N=64$. As summarized in Table~\ref{tab:tiling_sensitivity}, normalized throughput varies by at most 3\% across the tested values of $M$.
This weak dependence on $M$ suggests that our default configuration ($M=128$) is near-optimal and robust across a wide range of document lengths.

\begin{table}[t]
\centering
\caption{Normalized throughput vs. prefilling tile size $M$ (fixed $N=64$).}
\label{tab:tiling_sensitivity}
\begin{tabular}{lccc}
\toprule
Doc Length (tokens) & M = 64 & M = 128 (Ours) & M = 256 \\
\midrule
Short (256) & 1.00$\times$ & 0.99$\times$ & 0.97$\times$ \\
Medium (4K) & 0.99$\times$ & 1.00$\times$ & 0.99$\times$ \\
Long (16K) & 0.98$\times$ & 1.00$\times$ & 1.00$\times$ \\
\bottomrule
\end{tabular}
\end{table}



\blue{
\subsection{Beyond RAG: Few-Shot Classification}\label{appx:beyond_rag}
To demonstrate applicability beyond RAG, we evaluate \ourmethod on a non-RAG few-shot classification workload built from AG News~\citep{agnews2015}. Exemplar chunks recur across requests at different positions and in different orders. No RAG-like document-based retrieval is performed; instead, recurring chunks are detected based on exact token match.
As shown in Table~\ref{tab:agnews}, \ourmethod achieves a 1.31$\times$ TTFT speedup over Block-Attention in the skewed setting. Even in the uniform setting, \ourmethod delivers a 1.05$\times$ TTFT speedup, confirming that our method benefits any workload with recurring text chunks.

\begin{table}[h]
\centering
\caption{Few-shot classification on AG News (GH200/H100).}
\label{tab:agnews}
\begin{tabular}{llcc}
\toprule
Setting & Method & TTFT Speedup \\
\midrule
Skewed & Block-Attn & 1.0$\times$  \\
Skewed & Lazy-Attn (ours) & {1.31$\times$} \\
\midrule
Uniform & Block-Attn & 1.0$\times$ \\
Uniform & Lazy-Attn (ours) & {1.05$\times$} \\
\bottomrule
\end{tabular}
\end{table}
}

\blue{
\subsection{Long-Context Memory-Constrained Setting}\label{appx:long_context_constrained}
We evaluate \ourmethod under a memory-constrained long-context setting to demonstrate substantial absolute gains. Batched requests are drawn from a shared pool of 16 documents (5 per request, $\sim$8K tokens/document) with independently shuffled orders, under a 10\,GB GPU KV pool on GH200/H100.

As shown in Table~\ref{tab:long_context_mem}, compared to Block-Attention (integrated into vLLM), \ourmethod reduces TTFT from 15.8\,s to 7.2\,s (2.2$\times$) and increases throughput from 0.42 to 0.80 req/s (1.9$\times$). It also improves the cache hit ratio from 9.6\% to 28.8\% (3.0$\times$) and reduces evictions/recomputations from 847 to 206 (4.1$\times$ fewer).

\begin{table}[h]
\centering
\caption{Long-context memory-constrained setting (16 docs $\times$ 8K tokens, 10\,GB KV pool).}
\label{tab:long_context_mem}
\begin{tabular}{lcccc}
\toprule
Method & TTFT (s) & Throughput & Hit Ratio & Evictions \\
\midrule
Block-Attention & 15.8 & 0.42 req/s & 9.6\% & 847 \\
Lazy-Attn (ours) & {7.2} & {0.80 req/s} & {28.8\%} & {206} \\
\bottomrule
\end{tabular}
\end{table}
}

\subsection{MEPIC-like Baseline Comparison}\label{appx:mepic}
Since an official MEPIC~\cite{mepic} implementation was not available, we implemented a MEPIC-like baseline matching its core design: fully NoPE KV storage with page-aligned positional handling, under the same engine, batching, and paging settings as \ourmethod. We evaluate on 100 requests sampled from a mixture of four datasets, with at most 5 concurrent requests. These results show that \ourmethod preserves near-baseline decoding latency (0.67\% vs.\ MEPIC's 16\% overhead) while eliminating extra KV preparation time. This is our reimplementation rather than the official MEPIC code.

\begin{table}[h]
\centering
\caption{Kernel-level comparison with MEPIC-like baseline.}
\label{tab:mepic_comparison}
\begin{tabular}{lcccc}
\toprule
System & Query-only & Decode & Extra KV \\
& prefill (ms) & latency (ms) & prep.\ (ms) \\
\midrule
vLLM (baseline) & 360.93 & 40.57 (+0\%) & 0 \\
EPIC & 361.77 & 40.61 (+0.10\%) & 41 \\
MEPIC-like & 376.48 & 47.19 (+16\%) & 23 \\
Lazy-Attn (ours) & {364.12} & {40.84 (+0.67\%)} & {0} \\
\bottomrule
\end{tabular}
\end{table}

\section{Use of Large Language Models}

We disclose that, for this paper, large language models were used to polish wording and grammar, including clarity and tone. LLMs were also used to assist with experimental workflows, including drafting utility scripts and reviewing code. The research ideas, algorithms, experimental design, figures, analyses, and conclusions were developed by the authors. All LLM-assisted outputs were manually reviewed, verified, and approved by the authors, who take full responsibility for the content of the paper.




\end{document}